%% file: main.tex
\documentclass{article} % For LaTeX2e
\usepackage{iclr2026/iclr2026_conference,times}

\input{iclr2026/math_commands.tex}

\usepackage{hyperref}
\usepackage{url}

\usepackage[utf8]{inputenc} % allow utf-8 input
\usepackage[T1]{fontenc}    % use 8-bit T1 fonts
\usepackage{hyperref}       % hyperlinks
\usepackage{url}            % simple URL typesetting
\usepackage{booktabs}       % professional-quality tables
\usepackage{amsfonts}       % blackboard math symbols
\usepackage{nicefrac}       % compact symbols for 1/2, etc.
\usepackage{microtype}      % microtypography
\usepackage{xcolor}         % colors

\usepackage{graphicx}
\usepackage{amsmath} 
\usepackage{lipsum}
\usepackage{xcolor}    % 颜色支持
\usepackage{booktabs}  % 专业表格样式
\usepackage{array}     % 列宽调整支持
\usepackage{tcolorbox}
\usepackage{colortbl}
\usepackage[table]{xcolor}
\usepackage{array}
\usepackage{booktabs}
\usepackage{subcaption}

\definecolor{tableHeader}{rgb}{.8, .8,  1}
\definecolor{tableRow}{RGB}{229,240,255}
\usepackage{multirow}  % 多行合并支持
\usepackage{wrapfig}  % 必须加载
\usepackage{pifont}       % \ding{xx}
\usepackage{bbding}       % \Checkmark,\XSolid,... (需要和pifont宏包共同使用)
\usepackage{fontawesome}
\usepackage{pifont}       % \ding{xx}
\usepackage{bbding}       % \Checkmark,\XSolid,... (需要和pifont宏包共同使用)
\usepackage{amsmath} % For \text and \exp

\usepackage{amsmath, amssymb} % For math symbols
\usepackage{algorithm}      % For algorithm environment
\usepackage{algpseudocode}  % For pseudocode layout

\usepackage{amsmath, amssymb} % For math symbols
\usepackage{algorithm}
\usepackage{booktabs}
\usepackage{multirow}
\usepackage{siunitx}
\usepackage[table]{xcolor}
\usepackage{caption}
\usepackage{ulem}
\usepackage{enumitem}

\usepackage{makecell}
\title{MOSS-ChatV: Reinforcement Learning with Process Reasoning Reward for Video Temporal Reasoning}

\iclrfinalcopy
\author{
  \textbf{Sicheng Tao}\textsuperscript{1}\thanks{Core contribution.},
  \textbf{Jungang Li}\textsuperscript{1,2}\footnotemark[1],
  \textbf{Yibo Yan}\textsuperscript{1,2}\footnotemark[1],
  \textbf{Junyan Zhang}\textsuperscript{1},
  \textbf{Yubo Gao}\textsuperscript{1},
  \textbf{Hanqian Li}\textsuperscript{1}\\
  \textbf{ShuHang Xun}\textsuperscript{3},
  \textbf{Yuxuan Fan}\textsuperscript{1},
  \textbf{Hong Chen}\textsuperscript{1,2},
  \textbf{Jianxiang He}\textsuperscript{1},
  \textbf{Xuming Hu}\textsuperscript{1,2}\thanks{Correspondence: \texttt{xuminghu97@gmail.com}}
  \\[2pt]
  \textsuperscript{1}\,HKUST (GZ) \quad
  \textsuperscript{2}\,HKUST \quad
  \textsuperscript{3}\,HIT
}

\begin{document}

\maketitle

\input{preamble}
\input{sections/0_abstract}

\input{sections/1_introduction}
\input{sections/2_video_pred_task}
\input{sections/3_mosschatv}
\input{sections/4_experiment}

\input{sections/5_related_work}
\input{sections/6_conclusion}



\bibliography{main}
\bibliographystyle{iclr2026/iclr2026_conference}
\newpage

\appendix
\input{sections/appendix}
% \section{Appendix}

\end{document}

%% file: iclr2026/math_commands.tex
\usepackage{amsmath,amsfonts,bm}

\def\eqref#1{equation~\ref{#1}}
\def\1{\bm{1}}

\DeclareMathAlphabet{\mathsfit}{\encodingdefault}{\sfdefault}{m}{sl}
\SetMathAlphabet{\mathsfit}{bold}{\encodingdefault}{\sfdefault}{bx}{n}

%% file: preamble.tex
% \usepackage{pifont} % For \xmark and \cmark
% --- inline annotations
%
\newcommand{\red}[1]{{\color{red}#1}}
\newcommand{\todo}[1]{{\color{red}#1}}
\newcommand{\TODO}[1]{\textbf{\color{red}[TODO: #1]}}
% \newcommand{\OurBench}{RTV-Bench}
% \newcommand{\OurModel}{RTV-Bench}
% --- disable by uncommenting  
% \renewcommand{\TODO}[1]{}
% \renewcommand{\todo}[1]{#1}

\definecolor{my_green}{RGB}{51,102,0}
\definecolor{my_red}{RGB}{204, 0, 0}
\definecolor{api}{HTML}{ECF4FF}
\definecolor{correct_answer}{HTML}{DCF2DC}

\definecolor{cljg}{HTML}{0E7468}
\newcommand{\ljg}[1]{\color{red}{{ljg:#1}}}

\newcommand{\cmark}{\textcolor{my_green}{\ding{51}}} % ✔
\newcommand{\xmark}{\textcolor{my_red}{\ding{55}}} % ✘
\newcommand{\eg}{\textit{e.g.,\ }}

\newcommand{\NAME}{\textsc{VidComposition}\xspace}

\definecolor{myblue}{HTML}{EBEBFF}

%% file: sections/0_abstract.tex
\begin{abstract}

Video reasoning has emerged as a critical capability for multimodal large language models (MLLMs), requiring models to move beyond static perception toward coherent understanding of temporal dynamics in complex scenes. Yet existing MLLMs often exhibit \textbf{process inconsistency}, where intermediate reasoning drifts from video dynamics even when the final answer is correct, undermining interpretability and robustness.
To address this issue, we introduce \textbf{MOSS-ChatV}, a reinforcement learning framework with \textbf{a Dynamic Time Warping (DTW)–based process reward}. This rule-based reward aligns reasoning traces with temporally grounded references, enabling efficient process supervision without auxiliary reward models. We further identify dynamic state prediction as a key measure of video reasoning and construct \textbf{MOSS-Video}, a benchmark with annotated reasoning traces, where the training split is used to fine-tune MOSS-ChatV and the held-out split is reserved for evaluation.
MOSS-ChatV achieves 87.2\% on the MOSS-Video (test) and improves performance on general video benchmarks such as MVBench and MMVU. The framework consistently yields gains across different architectures, including Qwen2.5-VL and Phi2, confirming its broad applicability. Evaluations with GPT-4o-as-judge further show that MOSS-ChatV produces more consistent and stable reasoning traces.

\end{abstract}

%% file: sections/1_introduction.tex
\section{Introduction}
\label{sec:1_introduction}
Multimodal Large Language Models (MLLMs) have shown remarkable progress in vision–language tasks such as image captioning, visual question answering, and video description~\citep{cheng2024videollama, zhang2025videollama, liang2024survey, caffagni2024revolution}. Extending these advances from images to videos has attracted great attention, as videos contain richer temporal and causal information. However, video reasoning---requiring models to connect visual observations with temporal dynamics and causal dependencies---remains particularly challenging for current MLLMs.

Existing Video-MLLMs are predominantly trained through supervised fine-tuning on large-scale video–text pairs~\citep{li2024llava}. While effective for basic understanding, this paradigm leaves models weak in reasoning-intensive tasks. A fundamental issue is the scarcity of datasets that provide fine-grained temporal reasoning supervision. Yet, videos inherently encode dense supervisory signals in their temporal evolution. The core challenge lies in exploiting these temporal signals to strengthen reasoning: models must not only recognize the present state but also infer future trajectories from context and world knowledge. Prior work such as VoT~\citep{fei2024videoofthoughtstepbystepvideoreasoning} has shown the close coupling between video prediction and reasoning, underscoring that temporal state prediction can serve as a proxy for reasoning ability. To operationalize this insight, we construct \textbf{MOSS-Video}, a dataset for video state prediction with annotated reasoning traces. The dataset is partitioned into training and test splits, enabling process-supervised learning while ensuring held-out evaluation.

Reinforcement learning (RL) offers a promising path for strengthening reasoning in MLLMs. However, recent studies (\eg, Video-UTR~\citep{yu2025unhackable}) reveal a ``temporal hacking'' problem, where models bypass temporal reasoning and directly guess outcomes. This highlights the necessity of explicit process-level supervision. RL with process feedback has proven effective in domains such as mathematics and code generation~\citep{shao2024deepseekmathpushinglimitsmathematical, ye2025processsupervisedreinforcementlearningcode}. Motivated by this, we design a rule-based \textbf{Process Reasoning Reward (PRR)} for video reasoning. Specifically, we employ a two-stage ``split-align'' strategy: (1) decomposing reasoning traces into sequential substeps, and (2) aligning generated and reference processes via subsequence Dynamic Time Warping (DTW). The resulting alignment distance provides a reward signal that supervises temporal coherence without the need for a learned reward model. Leveraging PRR together with the MOSS-Video training split, we fine-tune \textbf{MOSS-ChatV} using GRPO~\citep{deepseekr1}, as illustrated in Figure~\ref{fig:3_overall_pipeline}.

Extensive experiments validate the effectiveness of our approach. See figure~\ref{fig:1_reasoning_and_prediction_across_different_mllm} for the case demonstrations. \textbf{MOSS-ChatV} achieves 87.2\% accuracy on the MOSS-Video test set, surpassing strong closed-source baselines such as GPT-4o. It also improves general video understanding, reaching 67.6\% on MVBench~\citep{li2024mvbench}, and performs competitively on real-time benchmarks such as RTVBench~\citep{xun2025rtv}. Moreover, the framework consistently boosts reasoning quality across architectures including Qwen2.5-VL and TinyLLaVA-Video. Automatic evaluation with GPT-4o as a judge further shows that MOSS-ChatV produces more consistent and stable reasoning traces. Our main contributions are as follows:
\begin{itemize}[leftmargin=15pt]
    \item We construct \textbf{MOSS-Video}, a video state prediction dataset with reasoning annotations, split into training and test partitions for process-supervised reinforcement learning and held-out evaluation.
    \item We propose a rule-based \textbf{Process Reasoning Reward (PRR)} based on subsequence DTW and integrate it into a reinforcement learning framework, \textbf{MOSS-ChatV}, trained with GRPO. This design enables efficient temporal alignment and process supervision without training additional reward models.
    \item Through extensive experiments, we demonstrate that MOSS-ChatV achieves state-of-the-art performance on the MOSS-Video (test), improves general video understanding benchmarks such as MVBench and MMVU, and yields consistent gains across different architectures including Qwen2.5-VL and TinyLLaVA-Video.
\end{itemize}

%fig:1_reasoning_and_prediction_across_different_mllm
\input{figures/1_reasoning_and_prediction_across_different_mllm}
% 为了解决这些挑战，我们提出了MOSS-ChatV模型。MOSS-ChatV在强化学习的框架中引入动态时间规整，旨在更好地对齐视频帧与文本描述，从而提升时序推理的能力。

%% file: figures/1_reasoning_and_prediction_across_different_mllm.tex
\begin{figure*}[tbp]
    \centering
    \includegraphics[width=1.0\linewidth]{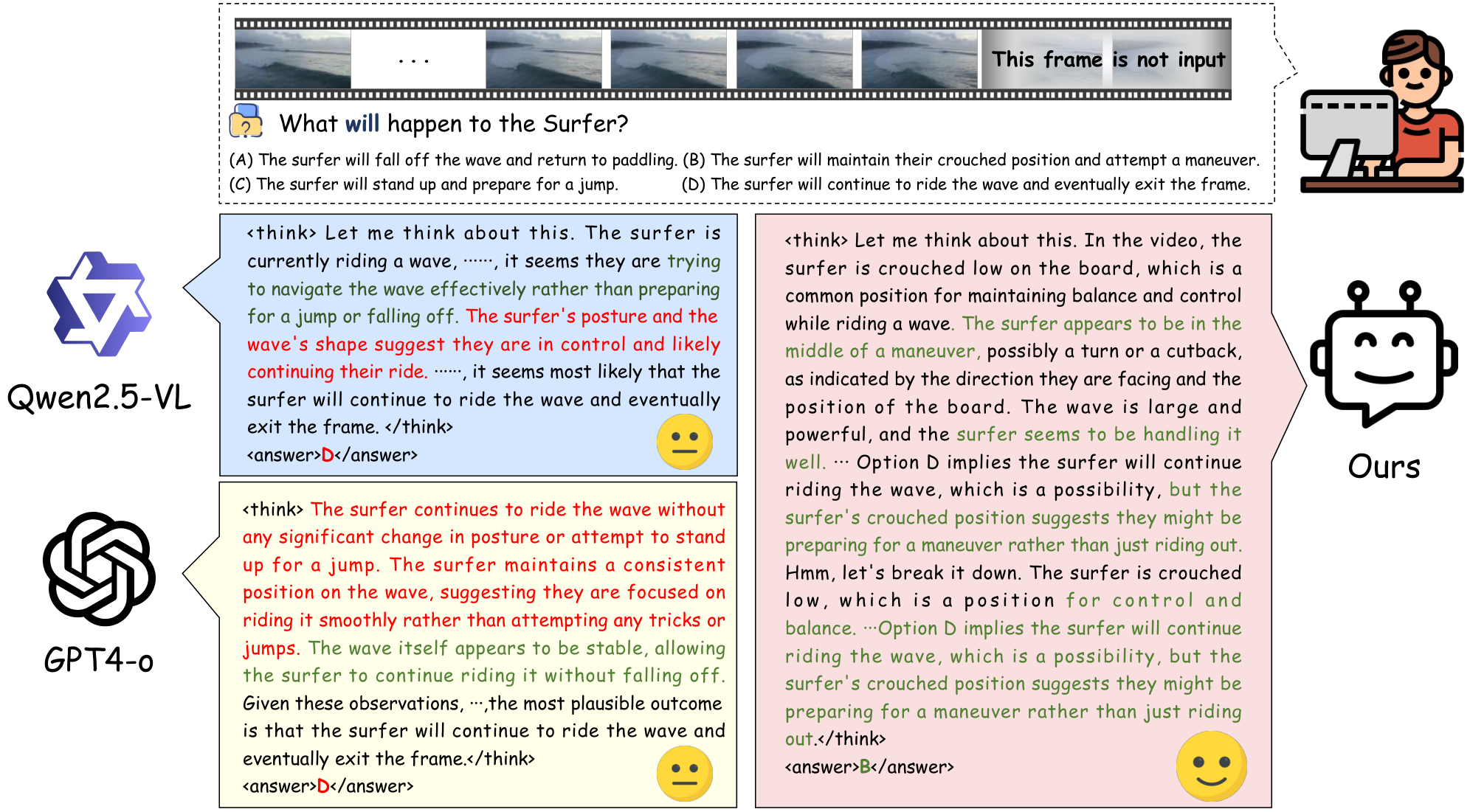}
    %% 不同模型在推理任务上的表现，说明现象，指出MOSS-ChatV的优势
    % 这个图展示了不同模型在视频状态预测任务的上的响应，绿色text代表推理正确的关键点，红色text代表推理错误的关键点。可以发现，相比于其他模型，MOSS-ChatV捕获到了更细粒度的状态（the surfer's couched position），并且正确的将这个状态进行外推（preparing for a mancuver) 从而实现了更加连贯正确的推理.
    \caption{Illustration of the responses across different models on the video state prediction task, where \textcolor[RGB]{42,148,127}{\textbf{green}} text indicates correctly reasoned key points and \textcolor[RGB]{178,57,88}{\textbf{red}} text denotes reasoning errors. Comparative analysis reveals that MOSS-ChatV captures more fine-grained states (\textit{e.g.}, the surfer’s crouched position) compared to other models. Crucially, it accurately extrapolates this state (preparing for a maneuver), thereby achieving more coherent and correct reasoning.}
    \label{fig:1_reasoning_and_prediction_across_different_mllm}
\end{figure*}

%% file: sections/2_video_pred_task.tex
\section{Preliminary}
\label{sec:preliminary}

\subsection{Video State Prediction and Reasoning}
\label{subsec:vsp}
We consider video state prediction as follows: given a video $V$ and a query $q$ specifying a target object, the model must (i) identify the object, (ii) infer its current or imminent state, and (iii) provide a temporally grounded explanation. An illustrative example is shown in Figure~\ref{fig:1_reasoning_and_prediction_across_different_mllm}. 
VoT~\citep{fei2024videoofthoughtstepbystepvideoreasoning} demonstrates that decomposing the task via Chain-of-Thought (CoT)---including task definition, object recognition/tracking, behavior analysis, answer ranking, and verification---yields a human-like reasoning path and highlights the tight coupling between prediction and reasoning. 
Different from the prompt-based paradigm in VoT, our approach learns this capability via reinforcement learning with a process-level reward, integrating temporal reasoning into the model’s latent space to enable end-to-end reasoning and prediction.

\subsection{Group Relative Policy Optimization (GRPO)}
\label{subsec:grpo}
Recent work (DeepSeek-R1)~\citep{deepseekr1} introduced Group Relative Policy Optimization (GRPO), which has spurred effective adaptations for multimodal LLMs~\citep{videor1,li2025videochat,wang2025timezero,zhang2025tinyllava}. 
At a high level, for each input, GRPO samples a group of $G$ candidate responses from the current policy $\pi_\theta$, compares their relative performance via a scalar reward, and updates the policy without learning a value function. 
We adopt GRPO as our optimization backbone due to its simplicity and strong empirical stability.

\paragraph{Notation.}
For one input, let the sampled response set be $\mathcal{O}=\{o_i\}_{i=1}^{G}$ with corresponding scalar rewards $\{\mathcal{R}_i\}_{i=1}^{G}$. 
GRPO computes a standardized advantage for each response:
\begin{equation}
A_i \;=\; \frac{\mathcal{R}_i - \mu}{\sigma}\,, 
\qquad 
\mu \;=\; \operatorname{mean}\!\big(\{\mathcal{R}_i\}_{i=1}^{G}\big),\quad 
\sigma \;=\; \operatorname{std}\!\big(\{\mathcal{R}_i\}_{i=1}^{G}\big).
\label{eq:adv}
\end{equation}
The learning objective encourages higher-advantage responses under importance weighting, while regularizing the policy against a fixed reference policy $\pi_{\text{ref}}$:
\begin{equation}
\begin{split}
    \mathcal{L}_{\mathrm{GRPO}}(\theta) \;=\; & \mathbb{E}_{\mathbf{o} \sim (\pi_{\theta}^{\mathrm{old}})}
    \left[
    \frac{1}{G} \sum_{i=1}^{G}
    \min\left(
    \frac{\pi_{\theta}(o_i)}{\pi_{\theta}^{\mathrm{old}}(o_i)} A_i,
    \;
    \text{clip}\left(\frac{\pi_{\theta}(o_i)}{\pi_{\theta}^{\mathrm{old}}(o_i)}, 1-\epsilon, 1+\epsilon\right) A_i
    \right)
    \right] \\
    & \;-\; \beta\, D_{\mathrm{KL}}\!\big(\pi_{\theta}\,\big\|\,\pi_{\text{ref}}\big)
\end{split}
\label{eq:grpo_obj}
\end{equation}
Here $\pi_{\theta}^{\mathrm{old}}$ denotes the behavior policy used for sampling the group, $\epsilon$ denotes the range of the clip operation, and $D_{\mathrm{KL}}(\cdot\|\cdot)$ is the Kullback--Leibler divergence. The importance ratio reweights each response $o_i$ to correct for the sampling distribution, while the KL term (scaled by $\beta>0$) controls policy drift.

\paragraph{Accuracy Reward.}
For multiple-choice or short-answer settings, a binary accuracy signal provides a simple yet effective supervision:
\begin{equation}
\mathcal{R}_{\mathrm{acc}}(a_{\mathrm{model}}, a_{\mathrm{gt}}) \;=\;
\begin{cases}
1, & \text{if } a_{\mathrm{model}} = a_{\mathrm{gt}},\\[2pt]
0, & \text{otherwise.}
\end{cases}
\label{eq:accuracy_reward}
\end{equation}

\paragraph{Format Reward.}
In many applications, outputs must follow a specified schema (\eg, \texttt{<think>...</think><answer>...</answer>}) to expose intermediate reasoning. Let $o_{\mathrm{model}}$ denote the full model output and $\mathcal{F}$ the required format:
\begin{equation}
\mathcal{R}_{\mathrm{fmt}}(o_{\mathrm{model}}, \mathcal{F}) \;=\;
\begin{cases}
1, & \text{if } o_{\mathrm{model}} \text{ adheres to } \mathcal{F},\\[2pt]
0, & \text{otherwise.}
\end{cases}
\label{eq:format_reward}
\end{equation}

\noindent
Accuracy and format rewards are effective foundations for RL fine-tuning, but they do not explicitly supervise temporal logic. In our method (Section~\ref{sec:3_method}), we therefore introduce a process-level reward to align intermediate reasoning with reference temporal processes, complementing $\mathcal{R}_{\mathrm{acc}}$ and $\mathcal{R}_{\mathrm{fmt}}$ within the GRPO framework. Algorithm~\ref{grpo} summarizes the overall optimization steps.

%% file: sections/3_mosschatv.tex
\section{Process Reasoning Reward}
\label{sec:3_method}
%% figure_3_overall_pipeline
\input{figures/3_overall_pipeline}

Addressing the limitations of conventional rewards in guiding complex temporal reasoning, we introduce a Process Reasoning Reward (PRR), denoted as $\mathcal{R}_{\text{proc}}$. This reward leverages reference annotations embodying an ideal 'gold standard' reasoning process. Crucially, this mechanism achieves nuanced process supervision by effectively leveraging efficient, robust algorithms, avoiding the need for potentially complex or computationally expensive large model-based evaluators.

\paragraph{Reasoning Step Serialization} The first step is segmentation for reasoning texts. The model's intermediate reasoning (\textit{e.g.}, content within \texttt{<think>...</think>} tags) and the reference counterpart are segmented into sequences of textual steps using NLP tools (\textit{e.g.}, nltk library). Though not affecting overall temporal information, this segmentation enables finer-grained analysis in the next step by splitting long texts into sequences. 
%algorithm:3_GRPO_for_one_training_sample
\input{algorithm/3_GRPO_for_one_training_sample}

Let $T_{gen}$ represent the intermediate reasoning content generated by the model, and $T_{ref}$ represent the reference reasoning content. These are segmented into sequences of textual steps using NLP tools (denoted as $\mathcal{N}$):
\begin{equation}
Seq_{gen} = \{g_1, \dots, g_m\} = \mathcal{N}(T_{gen})
\end{equation}
\begin{equation}
Seq_{ref} = \{r_1, \dots, r_n\} = \mathcal{N}(T_{ref})
\end{equation}

\paragraph{Temporal Alignment via Subsequence DTW}
For the second step, We employ Subsequence Dynamic Time Warping (SDTW), detailed in Algorithm~\ref{sdtw}, a highly efficient dynamic programming algorithm, to quantify the alignment between two sequences with different lengths. SDTW optimally identifies the best-matching subsequence within the model's reasoning sequence ($Seq_{gen}$) that corresponds to the entire reference sequence ($Seq_{ref}$), by minimizing a cumulative distance.
%% algorithm:3_subsequence_DTW
\input{algorithm/3_subsequence_DTW}
This cumulative distance, minimized by SDTW, is built upon the pairwise distances $d(g_j, r_i)$ between an individual generated step $g_j \in Seq_{gen}$ and the annotated reference step $r_i \in Seq_{ref}$.
To define $d(g_j, r_i)$, our goal is to comprehensively yet efficiently measure the textual similarity between these steps. This is achieved by leveraging several rule-based ROUGE scores. We use ROUGE-1 and ROUGE-2 to capture n-gram overlap between $g_j$ and $r_i$. To evaluate sequence-level structural similarity, ROUGE-L is used for preserving the sentence-internal logical order within each step.

The distance $d(g_j, r_i)$ is then formally defined as one minus the average of average of these ROUGE scores:
\begin{equation}
\text{ROUGE}_{\text{avg}}(g_j, r_i) = \frac{\text{ROUGE-1}(g_j, r_i) + \text{ROUGE-2}(g_j, r_i) + \text{ROUGE-L}(g_j, r_i)}{3}
\end{equation}
\begin{equation}
d(g_j, r_i) = 1 - \text{ROUGE}_{\text{avg}}(g_j, r_i)
\end{equation}
The minimum cumulative distance is then defined as:
\begin{equation}
D_{sdtw} = SUBSEQUENCE\_DTW(D)
\end{equation}

We adopt Subsequence Dynamic Time Warping (SDTW) for its ability to align a reference reasoning path ($Seq_{ref}$) within potentially longer model-generated sequences ($Seq_{gen}$), enabling process supervision with explicit temporal signals. A key advantage is SDTW's compatibility with reinforcement learning: it avoids penalizing exploratory segments outside the optimal alignment while still rewarding correct paths. The algorithm provides tunable alignment strictness through parameters like \textit{jump steps} (Algorithm~\ref{sdtw}, figure~\ref{fig:3_overall_pipeline}), permitting controlled tolerance for minor deviations in the reasoning trajectory. This balance of flexibility and precision makes SDTW ideal for guiding reasoning processes without stifling exploration.

% An appendix provides a comparative illustration. % (If you still want to mention the appendix)

% A key feature of this `subsequence' approach is that it does not penalize exploratory steps or additional content within $\text{Seq}_{gen}$ that fall outside the optimal alignment with $\text{Seq}_{ref}$. 

\paragraph{Distance-to-Reward Transformation} The final minimum cumulative distance $D_{sdtw}$ from SDTW is transformed into the reward value $\mathcal{R}_{\text{proc}}$ via a transformation function $\mathcal{T}$:
\begin{equation}
\mathcal{R}_{\text{proc}} = \mathcal{T}(D_{sdtw})
\end{equation}
\begin{equation}
\mathcal{T}(D_{sdtw}) = \exp(-\alpha \cdot D_{sdtw})
\end{equation}
where $\alpha > 0$ is a tunable hyperparameter that controls the sensitivity or decay rate of the reward with respect to the distance.

Then we can get the total reward $\mathcal{R}_{\text{total}, i}$ for the $i$-th response in the sampled group of responses, by combining its specific process reward $\mathcal{R}_{\text{proc}, i}$ with its accuracy $\mathcal{R}_{\text{acc}, i}$ and format $\mathcal{R}_{\text{fmt}, i}$:
\begin{equation}
\mathcal{R}_{\text{total}, i} = \mathcal{R}_{\text{proc}, i} + \mathcal{R}_{\text{acc}, i} + \mathcal{R}_{\text{fmt}, i}
\end{equation}
This $\mathcal{R}_{\text{total}, i}$ corresponds to the $\mathcal{R}_i$ used in the GRPO advantage calculation (Equation \ref{eq.adv}) for the $i$-th response within the group $\{o_1, \dots, o_G\}$.

Consequently, the resulting $\mathcal{R}_{proc}$ provides a computationally efficient yet powerful reward signal for reinforcement learning. It uniquely encourages temporal coherence in reasoning, validates the inclusion and ordering of essential logical steps, and maintains sensitivity to the relevance of generated content, thereby offering comprehensive guidance towards generating both accurate and logically sound reasoning processes.

\paragraph{MOSS-Video Dataset}To support process-supervised reinforcement learning, we construct \textbf{MOSS-Video}, a large-scale video state prediction dataset derived from ShareGPT4Video~\citep{chen2024sharegpt4video}. Each sample is annotated with object states and corresponding reasoning traces, enabling models to predict future states conditioned on visual context. The dataset is partitioned into a training split (11,654 samples, 1,218 unique videos) and a held-out test split (2,836 samples, 479 unique videos). Basic statistics are summarized in Table~\ref{tab:3_comparison}, including average video length and annotation span. Annotation pipelines and further details are provided in Appendix~\ref{sec:appendix_details_of_moss_video}.

%% tab:3_comparison
\input{tables/3_comparison}

%% file: figures/3_overall_pipeline.tex
\begin{figure*}
    \centering
    \includegraphics[width=1\linewidth]{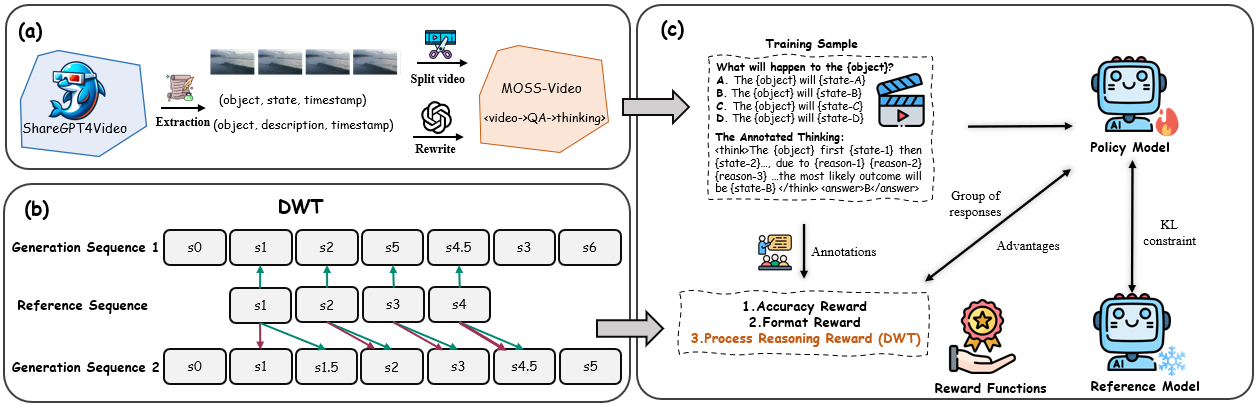}
    %这个图展示了本研究的核心点。（a）我们利用了细致标注的sharegpt4video数据去生成MOSS-Video数据集，基于其良好的多层次多时间戳的标注信息，我们可以用gpt4.1生成符合我们要求的数据，最终我们丢去靠后的一些时间戳标注和视频帧（这部分应该是预测的答案）剩下的就构成了输入给模型做预测的视频帧了。（b）展现了两个关于subsequence dtw的核心cases，“s" stands for a statement,箭头的粗细代表着匹配后的距离，越粗距离越远,generation sequence 1展现了一个错序的匹配，可以看到s3->s5和s4->s4.5有较高的距离。generation sequence 2展现了jump step的用法，当jump step设为2时（），在匹配最短距离路径的时候可以跳过一些中间序列，从而得到更小的累积距离。（c）展现了GRPO的训练流程
    \caption{Overall training pipeline of MOSS-ChatV. 
(a) Construction of the MOSS-Video dataset from ShareGPT4Video with multi-level temporal annotations, where future states are masked as prediction targets. 
(b) Subsequence DTW alignment: green dashed lines denote strict sequential matching, while red solid lines allow jumps (\textit{jump step}=2) to reduce cumulative distance. 
(c) GRPO workflow integrating accuracy, format, and process rewards.}

    \label{fig:3_overall_pipeline}
\end{figure*}

%% file: algorithm/3_GRPO_for_one_training_sample.tex
\begin{algorithm}[t]
\caption{GRPO with Process Reasoning Reward (PRR) for one training sample}
\label{grpo}
\begin{algorithmic}[1]
\Require Training sample $(V, Q, a_{\text{gt}}, T_{\text{ref}})$; 
Policy model $M_{\text{policy}}(\pi_\theta)$; 
Reference model $M_{\text{ref}}(\pi_{\text{ref}})$; 
Format $\mathcal{F}$
\State Sample $G$ candidate outputs from policy: 
$\mathcal{O} = \{o_i=(T_{\text{gen}}, a_{\text{model}})\}_{i=1}^G$
\State Initialize reward list $\mathcal{R} = [\ ]$
\For{each $o_i \in \mathcal{O}$}
    \State Segment reference: $Seq_{\text{ref}} = \{r_1,\dots,r_n\} \gets \mathcal{N}(T_{\text{ref}})$
    \State Segment generation: $Seq_{\text{gen}} = \{g_1,\dots,g_m\} \gets \mathcal{N}(T_{\text{gen}})$
    \State Build distance matrix $\mathbf{D} \in \mathbb{R}^{m \times n}$ with
    \[
        D_{j,k} = 1 - \text{ROUGE}_{\text{avg}}(g_j, r_k), 
        \quad j\in[1,m],\ k\in[1,n].
    \]
    \State Compute $D_{\text{sdtw}} \gets \text{SUBSEQUENCE\_DTW}(\mathbf{D})$
    \State Process reward: $\mathcal{R}_{\text{proc}} = \exp(-\alpha \cdot D_{\text{sdtw}})$
    \State Total reward: 
    \[
      \mathcal{R}_i = 
      \mathcal{R}_{\text{acc}}(a_{\text{model}}, a_{\text{gt}})
      + \mathcal{R}_{\text{fmt}}(o_i, \mathcal{F})
      + \mathcal{R}_{\text{proc}}
    \]
    \State Append $\mathcal{R}_i$ to $\mathcal{R}$
\EndFor
\State Standardize advantages:
\[
A_i = \frac{\mathcal{R}_i - \mu}{\sigma}, \quad
\mu = \operatorname{mean}(\mathcal{R}), \ \sigma = \operatorname{std}(\mathcal{R})
\]
\State Compute GRPO objective with clipping:
\begin{equation}
\begin{split}
    \mathcal{L}_{\mathrm{GRPO}}(\theta) \;=\; & 
    \mathbb{E}_{\mathbf{o} \sim \pi_{\theta}^{\mathrm{old}}}
    \left[
    \frac{1}{G} \sum_{i=1}^{G}
    \min\Bigg(
    \frac{\pi_{\theta}(o_i)}{\pi_{\theta}^{\mathrm{old}}(o_i)} A_i,\;
    \text{clip}\!\left(\frac{\pi_{\theta}(o_i)}{\pi_{\theta}^{\mathrm{old}}(o_i)},\, 1-\epsilon,\, 1+\epsilon\right) A_i
    \Bigg)
    \right] \\
    & - \beta\, D_{\mathrm{KL}}\!\big(\pi_{\theta}\,\big\|\,\pi_{\text{ref}}\big).
\end{split}
\label{eq:grpo_obj}
\end{equation}
\State Update policy: $M_{\text{policy}}.\text{update}(\mathcal{L}_{\text{GRPO}})$
\end{algorithmic}
\end{algorithm}

%% file: algorithm/3_subsequence_DTW.tex
\begin{algorithm}
    \caption{Subsequence DTW}
    \label{sdtw}
    \begin{algorithmic}[1]
        \Function{Subsequence\_DTW}{$\mathbf{D}, k_{\text{ref}}, k_{\text{target}}$}
            \Comment{$\mathbf{D}$: Cost matrix ($n \times m$), $k_{\text{ref}}$: max reference jump, $k_{\text{target}}$: max target jump}
            \State Initialize $\mathcal{P} \in \mathbb{R}^{(n+1) \times (m+1)}$ with $\mathcal{P}[0,j] \gets 0$ for $j \in [0,m]$, $\mathcal{P}[i,0] \gets \infty$ for $i \in [1,n]$
            \For{$i \gets 1$ to $n$}
                \For{$j \gets 1$ to $m$}
                    \State $\text{diag\_cost} \gets \mathcal{P}[i-1,j-1]$ \Comment{Match current points (diagonal move)}
                    \State $\text{up\_cost} \gets \min_{1 \leq k \leq \min(k_{\text{ref}},i)} \mathcal{P}[i-k,j]$ \Comment{Skip $k$ points in reference sequence (vertical move)}
                    \State $\text{left\_cost} \gets \min_{1 \leq k \leq \min(k_{\text{target}},j)} \mathcal{P}[i,j-k]$ \Comment{Skip $k$ points in target sequence (horizontal move)}
                    \State $\mathcal{P}[i,j] \gets \mathbf{D}[i,j] + \min(\text{diag\_cost}, \text{up\_cost}, \text{left\_cost})$
                \EndFor
            \EndFor
            \State \Return $\min_{j \in [1,m]} \mathcal{P}[n,j]$ \Comment{Shortest distance to any endpoint in target sequence}
        \EndFunction
    \end{algorithmic}
\end{algorithm}

%% file: tables/3_comparison.tex
\begin{table}[h]
\centering
\caption{Comparison of MOSS-Video with representative video temporal reasoning datasets. 
Our dataset uniquely supports state prediction with explicit reasoning annotations.}
\label{tab:3_comparison}
\resizebox{\columnwidth}{!}{%
\begin{tabular}{lcccccc}
\toprule
Dataset & \#Samples & Avg. Video Len (s) & Understanding & Reasoning & Prediction \\
\midrule
ViTiB~\citep{vitcot}        & 1,382   & --     & \checkmark & \checkmark & $\times$ \\
NeXT-QA~\citep{nextqa}      & 3,870   & 40     & \checkmark & $\times$   & $\times$ \\
Video-R1-CoT-165K~\citep{videor1} & 116k & --     & $\times$   & \checkmark & $\times$ \\
\midrule
MOSS-Video (train) & 11,654  & 27.73 & \checkmark & \checkmark & \checkmark \\
MOSS-Video (test)  & 2,836   & 28.21 & \checkmark & \checkmark & \checkmark \\
\bottomrule
\end{tabular}
}
\end{table}

%% file: sections/4_experiment.tex
\section{Experiment}
\label{sec:experiment}

% 我们利用open-r1-video和video-r1提供的训练框架和moss_video_train数据直接对qwen2.5vl进行了强化微调。考虑实际的资源限制，我们选取了MVBench，Perception Test，tempcampass，video MME，RTV-Bench和MOSS-Video test 对MOSS-ChatV进行全面的评估（补充各个bench的细节）见表xxx,具体配置上我们采用的温度是0，多帧结果有标，分辨率约为448x224。
%表xx是一个具有代表性的cases。难度在于B选项和D选项的混淆，相较于笼统的ride the wave，B选项提供了更为详细的预测结论，这需要模型捕获到一些关键的状态，从而导向正确的预测。相较于其他模型，moss正是把握住了更多的状态，出了稳定动作之外，它也捕获到了动作前摇从而实现了连贯的动作预测

We directly performed reinforcement fine-tuning on the Qwen2.5VL model, leveraging the training frameworks provided by Open-R1-Video~\citep{wang-2025-open-r1-video} and Video-R1~\citep{videor1}, and utilizing the MOSS-Video train set. We selected a comprehensive suite of benchmarks for the holistic evaluation of MOSS-ChatV.
This suite includes MVBench~\citep{li2024mvbench}, TempCompass~\citep{liu2024tempcompassvideollmsreally}, Video MME~\citep{fu2024video}, RTV-Bench~\citep{xun2025rtv} and the MOSS-Video test set for our state prediction scenarios. These benchmarks collectively assess a wide range of video understanding capabilities, including temporal reasoning, action recognition, causal inference, and narrative comprehension. To demonstrate our method's generalizability, we further experiment on TinyLLaVA-Video~\citep{zhang2025tinyllava}, validating its effectiveness with a different language model (Phi2) and visual encoder (SigLIP).
The aggregated evaluation results are presented in Table~\ref{tab:main_results}.
Specific configurations for our evaluations included a sampling temperature of 0 to ensure deterministic outputs and an input video resolution of approximately 448x448 pixels. We tested our experiment on 4 NVIDIA A800 and trained MOSS-ChatV on 8 NVIDIA A800.

% Figure~\ref{tab:representative_case_analysis} presents a representative case study that highlights a common challenge in video-based state predictive reasoning, particularly in distinguishing between subtly different future outcomes. The primary difficulty in this illustrative instance arises from the potential for confusion between plausible options, specifically exemplified by the choice between option B and option D. In contrast to a more generalized or abstract prediction, such as "ride the wave" (which might characterize the nature of option D), option B offers a significantly more detailed and nuanced predictive conclusion regarding the subject's subsequent actions. Accurately selecting option B necessitates the model's ability to capture not only overt actions but also subtle yet critical preceding states or contextual cues within the video. These captured elements subsequently guide the model towards the correct, fine-grained prediction. Compared to other baseline models evaluated, our MOSS-ChatV model demonstrates a superior capability in discerning and integrating a broader range of these crucial states. Specifically, beyond recognizing stable, ongoing actions, MOSS-ChatV effectively captures anticipatory movements or preparatory actions. 

\subsection{Results}

% MOSS-Video test: 在MOSS-Video的测试集上 MOSS-ChatV取得了绝对的领先，这表明了我们的方法能很好地帮助模型学习到物品状态预测能力。此外，从qwen2.5vl到video-r1再到moss的性能提升可以看出，高质量的推理链条可以帮助模型更好的做到贴近视频内容的状态预测。

% MVBench 和 Temp 是专注于全方位时间理解的bench，moss提供了有竞争力的结果（MVBench上升，temp略微下降，总的来说是上升），体现了模型在时间理解能力上的进步。此外，两个bench都强调了视频的动态特性。因此，模型的进步意味着它能有效利用时间维度信息，而不是试图通过空间特征的“捷径”来解决问题，可以看出我们的数据和方法在解决捷径问题上提出了新的见解。

% RTV-Bench 是一个新颖的针对实时视频场景的benchmark，创新的使用了动态问答，既同一个问题的答案会随视频进展而发生变化。moss相较于qwen2.5vl有显著提升，说明模型对实时场景的处理能力也有进步，即使我们的训练数据中并不包含特定的视频流式数据

% Video MME是一个综合了长中短视频的全面的视频理解benchmark，较为全面地涵盖了视频理解的多个维度，moss依旧展现了有竞争力的性能

%我们的模型在各个维度的bench上都取得了良好的性能即使是在较为苛刻的测试环境中（16帧，448x224分辨率）。并且在横向的推理模型比较中，我们的模型质量也很有竞争力，这凸显了我们任务和数据。值得注意的是，我们在提升模型状态推理能力的时候只使用了单一任务，也就是状态预测任务，的训练数据，同时并没有丧失和遗忘qwen2.5vl自身已有高性能视频理解能力，这进一步说明了moss挖掘了更多模型本身的潜力

% Perception Test 待补充

% MOSS-ChatV 在MVBench， VideoMME，RTVBench和MOSS-Video test上，相较于basline，qwen2.5vl和同架构模型video-r1，都取得了最佳的表现，在Tempcompass上也取得了相较于基座模型的提升，值得注意的是，我们的方案仅仅使用视频预测任务这一单一任务数据集，但在通用视频bench都获得了提升，尤其是在MVBench和VideoMME这两个需要复杂推理的bench上，说明我们的方案深度挖掘了模型的潜力。% 从MOSS-Video test的结果可以看出来，推理行为对于视频预测任务是有促进作用，比如qwen2.5vl和video-r1都没有用MOSS-Video train进行训练但是在指标上有显著的提升。
% 这些结果可以说明视频预测任务确实对模型的推理能力有促进作用

%MOSS-ChatV性能超群，

In Table~\ref{tab:main_results}, our MOSS-ChatV model achieves state-of-the-art performance on MVBench, VideoMME, RTVBench, and MOSS-Video test compared to baseline models, Qwen2.5-VL, and the same architecture model Video-R1. It also demonstrates improvements over the Qwen2.5-VL on TempCompass.

\input{tables/4_main_performance}

The results from MOSS-Video test particularly indicate that reasoning capabilities contribute positively to video prediction tasks. Notably, while neither Qwen2.5-VL nor Video-R1 were trained on MOSS-Video Train data, but Video-R1 shows significant metric improvements, suggesting the benefits of reasoning.

%% figures:4_comparison
\input{figures/4_comparison}
We tested MOSS-Video using different input number of frames, result shown in Figure~\ref{fig:frame_quality}. The results demonstrate that increasing input frames enhances state prediction performance. MOSS-ChatV likely reaches peak accuracy with fewer frames due to its more efficient information extraction and reasoning capability.

It is worth emphasizing that our solution utilizes only a single-task dataset for video prediction, yet achieves performance gains across general video benchmarks. The improvements are especially pronounced on MVBench and VideoMME - both requiring complex reasoning - demonstrating that our approach effectively unlocks the model’s latent potential. These results collectively provide evidence that video prediction tasks indeed enhance models’ reasoning capabilities.
\vspace{-8pt}

\subsection{SDTW vs. DTW}
%Naive DTW和Subsequence DTW都可以对两个推理序列进行比对，但是由于Naive DTW会尝试将两个序列进行扭曲从而完全地匹配，也就是说会存在单个元素通过扭曲对应多个元素的情况，这会大大增加非等长序列的距离，这并不是我们所期望的，因为这会限制超出最小推理链条的合理探索。我们的实验表明，Naive DTW会导致奖励劫持，当模型输出的推理长度很短的时候，dtw的距离是最短的，模型只会输出一句无意义的推理，例如“<think>根据视频内容，正确答案是A</think><answer>A</answer>”。这个实验也说明仅仅把标注的推理过程当作金标准用于训练对模型提升效果有限，我们的策略仅将标注推理过程当做“最小”的金标准，在保证推理的质量的情况下，对模型的其他探索不作限制，从而更充分地挖掘了模型潜力。
\input{figures/5_response_len}
When comparing reasoning sequences, traditional Naive Dynamic Time Warping (DTW) and Subsequence DTW exhibit distinct behaviors. Naive DTW attempts to achieve a complete match between two sequences through warping, which can lead to a single element being mapped to multiple elements. This significantly inflates the distance metric for sequences of unequal length, a characteristic we find undesirable as it unduly penalizes valid model explorations that extend beyond the shortest annotated reasoning chain. Our experiments, Figure~\ref{fig:5_response_len}, demonstrate that Naive DTW can induce a ``reward hacking'' phenomenon: when the model outputs very short reasoning, the DTW distance is minimized, leading to trivial outputs like \texttt{<think>Based on the video content, the correct answer is A</think><answer>A</answer>}. This observation also highlights that treating annotated reasoning processes solely as an absolute ``gold standard'' for training offers limited benefits for model improvement. Consequently, our strategy positions annotated reasoning as a ``minimal'' gold standard. While ensuring the quality of reasoning, this approach avoids overly restricting the model's legitimate explorations beyond this baseline, thereby aiming to more comprehensively unlock and leverage the model's latent potential.

\subsection{Ablation Study}
% 我们在训练方法上进行了消融，训练数据均采用MOSS-Video train。分为MOSS-ChatV，MOSS no PPR和MOSS - sft。结果见表xx。可以发现，完整的带有Process Reasoning Reward的moss在各个bench上都体现出了绝对优势，对于moss no PRR来说，缺失的过程监督信号使推理过程可能会使推理变得更加随意，从而导致在时序表现上下降，说明增加时序的对齐信号对于视频任务有积极意义。从MOSS-ChatV with thinking reward，MOSS-ChatV without thinking reward和moss-sft的结果可以看出强化学习在视频任务上的优势，即使是缺失时序监督信号的moss no PRR依旧比带有时序监督信号的moss-sft有更好的性能

%% table:4_ablation
\input{tables/4_ablation}

We conduct ablation experiments using the MOSS-Video Train dataset, comparing three variants: MOSS-ChatV , MOSS-ChatV-no-PRR (MOSS-ChatV without process supervision), and supervised fine-tuned MOSS-ChatV-SFT. The results (see Table~\ref{table_ablation}) demonstrate that the complete MOSS-ChatV achieves superior performance across all benchmarks. The absence of process supervision in MOSS-ChatV-no-PRR leads to degraded temporal reasoning performance, confirming the importance of alignment signals for video understanding. Notably, even without temporal supervision, MOSS-ChatV-no-PRR outperforms MOSS-ChatV-SFT, highlighting the advantages of reinforcement learning over pure supervised training for video reasoning tasks.

\subsection{MLLM as a Judge for Reasoning Quality Evaluation}
% 在强化学习微调（RLHF）领域，尤其是在问答（QA）任务中，模型常表现出推理过程与最终答案不一致的问题【参考文献】。为探究此现象，我们采用GPT-4o作为评估者 (judge)，对模型输出的推理与答案进行多维度质量评估。该评估体系包含四个核心指标，GPT-4o对每个维度给评分（详细Prompt参见附录XX）。

% 具体指标如下：

% 推理-答案一致性 (0 或 1)：此为二元指标，若推理的最终结论与模型选定选项的内容一致则为1分，反之为0分。
% 推理内容重复度（0-10）：评估推理过程中是否存在冗余信息。重复度越高，得分越低，旨在衡量信息密度及避免潜在的偏见或错误放大。
% 逻辑连贯性与知识准确性（0-10）：直接评估推理过程的内在质量。逻辑越严谨、世界知识运用越准确，得分越高。
% 推理与视频内容关联度（0-10）：衡量推理是否紧密依据视频内容。关联度越高，得分越高，旨在避免无根据的臆测或与视频无关的联想。
% 我们从“moss test”和“video mme”的响应数据（详见表XX）中随机抽取2000条作为GPT-4o的评估对象，结果如表XX所示。研究发现：
% MOSS模型展现出最佳的推理-答案一致性。此优势可能源于其数据特性（例如，预测任务鼓励深度推理）及训练策略的协同作用（例如，推理过程监督与结果奖励相结合，促进了过程与结果的对齐，从而减少了潜在的奖励劫持风险，如为追求高响应长度而引入错误推理，或因训练集噪声被不当捕获）。
% Qwen2.5VL在推理内容简洁性（即低重复度）上表现最优，这或因其未经强化微调，从而保留了更广阔的探索空间。然而，这种自由探索也可能导致注意力分散，进而牺牲了部分推理-答案一致性。
% 基于“MOSS-Video”数据训练的模型（MOSS及MOSS-no-think版本）在逻辑连贯性与视频内容关联度上均获得较高评分。这凸显了状态预测任务数据在促进推理与视觉信息对齐、增强模型推理可靠性方面的价值。
% 综上，对推理内容的细致评估揭示了过程监督（process supervision）在强化微调中的潜在优势，为视频语言模型的优化提供了新的研究视角。

% tables:4_MLLM_as_a_judge
\input{tables/4_MLLM_as_a_Judge}
To investigate the quality of video reasoning texts, we employed GPT-4o as a judge to conduct a multi-dimensional quality assessment of the reasoning and answers generated by models. This assessment framework comprises four core metrics, for which GPT-4o assigns a score for each dimension (detailed dimension and prompts can be found in Appendix~\ref{fig:prompt_for_evaluating_reasoning}).

%强化微调中的过程监督对提升多个维度的推理质量有显著帮助。MOSS-ChatV比较均衡且整体优秀。和Video-R1相比，在差不多的一致性性能下，MOSS—ChatV有更高的信息密度，更通顺的逻辑和更高的视频相关性。和MOSS-ChatV no-PRR相比，MOSS—ChatV有更高的推理结果一致性，说明过程监督有效地引导了模型生成更可信的结果。虽然qwen2.5vl有最高的信息密度，但较低的其他指标暗示这可能是在进行无限制地思维发散，不利于生成高质量的推理内容
Process supervision within reinforcement fine-tuning demonstrates a significant contribution to enhancing the quality of reasoning across multiple dimensions.
MOSS-ChatV exhibits a well-balanced and overall excellent performance profile, shown in table~\ref{MLLM as a judge}.
Compared to Video-R1, while achieving comparable performance in Reasoning-Answer Consistency, MOSS-ChatV demonstrates higher information density (\textit{i.e.}, lower repetitiveness), more robust logical coherence, and greater relevance to video content.
Furthermore, when contrasted with its variant MOSS-ChatV-no-PRR, MOSS-ChatV achieves a higher degree of Reasoning-Answer Consistency. This suggests that process supervision effectively guides the model towards generating more credible and trustworthy outputs.
Although Qwen2.5-VL records the highest information density, its comparatively lower scores on other metrics imply that this conciseness might stem from unconstrained cognitive divergence, which could be detrimental to the generation of high-quality reasoning content.

%% file: tables/4_main_performance.tex
\begin{table*}[t]
  \centering
  \caption{Results of MOSS-ChatV and baselines on (a) general video understanding benchmarks and (b) video reasoning benchmarks. All results use 32-frame input setting. Our method consistently improves performance across both categories.}
  \label{tab:main_results}
  \begin{subtable}{\textwidth}
    \centering
    \small
    \caption{General Benchmarks}
    \begin{tabular}{lcccc}
      \toprule
      \textbf{Model} & \# \textbf{LLM} & MVBench & VideoMME & TempCompass \\
      \midrule
      Qwen2.5-VL~\cite{bai2025qwen25vltechnicalreport} & Qwen2.5-7B & \uline{67.1} & \uline{59.7} & 72.2 \\
      LLaVA-OneVision~\cite{li2024llava} & Qwen2-7B & 56.7 & 58.2 & -- \\
      TinyLLaVA-3B~\cite{zhang2025tinyllava} & Phi2-3B & 28.8 & 34.5 & 32.4 \\
      TinyLLaVA-3B + PRR & Phi2-3B & 29.0 & 35.1 & 45.1 \\
      Video-UTR~\cite{yu2025unhackable} & Qwen2-7B & 58.8 & 52.6 & 59.7 \\
      VideoChat-R1~\cite{li2025videochat} & Qwen2.5-7B & 66.2 & 58.8 & \uline{73.9} \\
      VideoChat-R1-thinking~\cite{li2025videochat} & Qwen2.5-7B & -- & 58.3 & \textbf{75.0} \\
      Video-R1~\cite{videor1} & Qwen2.5-7B & 63.9 & 59.3 & 73.2 \\
      \midrule
      \rowcolor{myblue}
      \textbf{MOSS-ChatV (ours)} & Qwen2.5-7B & \textbf{67.6} & \textbf{60.0} & 72.9 \\
      \bottomrule
    \end{tabular}
  \end{subtable}
  
  \vspace{0.5em} % small space between subtables

  \begin{subtable}{\textwidth}
    \centering
    \scriptsize
    \caption{Reasoning Benchmarks}
    \begin{tabular}{lcccccccc}
      \toprule
      \textbf{Model} & RTV-Bench & MOSS-Video$_{\text{test}}$ & MMVU$_{\text{mc}}$ & VideoMMMU & VCR-B$_{\text{mc}}$ & VSI-B$_{\text{mc}}$ & VSI-B$_{\text{reg}}$ \\
      \midrule
      Qwen2.5-VL & 32.8 & 67.0 & 60.0 & 48.1 & 33.7 & \uline{35.3} & 24.3 \\
      LLaVA-OneVision & 34.5 & 48.1 & -- & -- & -- & -- & -- \\
      TinyLLaVA-3B & -- & 65.9 & 39.0 & -- & -- & -- & -- \\
      TinyLLaVA-3B + PRR & -- & 82.5 & 40.3 & -- & -- & -- & -- \\
      Video-UTR & -- & 58.9 & -- & -- & -- & -- & -- \\
      VideoChat-R1 & -- & 70.8 & 62.7 & 50.0 & 34.5 & -- & -- \\
      VideoChat-R1-thinking & -- & 70.1 & 64.2 & 49.2 & \uline{35.3} & \textbf{35.9} & \uline{30.0} \\
      Video-R1 & \uline{46.5} & \uline{73.3} & \uline{64.8} & \textbf{52.3} & \textbf{38.4} & 30.8 & \textbf{39.7} \\
      \midrule
      \rowcolor{myblue}
      \textbf{MOSS-ChatV (ours)} & \textbf{46.6} & \textbf{86.6} & \textbf{66.2} & \uline{50.2} & \uline{35.3} & 35.2 & 28.2 \\
      \bottomrule
    \end{tabular}
  \end{subtable}
\end{table*}

%% file: figures/4_comparison.tex
% \begin{figure}[!htbp]
    % \centering
    % \begin{minipage}[t]{0.28\textwidth}
    %     \centering
    %     \includegraphics[width=\textwidth]{figures/assets/4_comparison_across_different_benchmarks.png}
    %     % 16帧 128分辨率 
    %     \caption{Visualization of the performance of various models on different evaluation sets.}
    %     \label{fig:combined_bit_accuracy}
        
    % \end{minipage}%
    % \hfill
%     \begin{minipage}[t]{0.45\textwidth}
%      \centering
%         \includegraphics[width=\textwidth]{figures/assets/4_comparison_across_frames.png}
%         % Video-R1和MOSS-ChatV在128分辨率，不同帧，在MOSS-Video$_{test}$上的性能表现
%         \caption{Performance impact of varying input frame counts.}
%         \label{fig:frames}   
%     \end{minipage}
%     \begin{minipage}[t]{0.45\textwidth}
%      \centering
%         \includegraphics[width=\textwidth]{figures/assets/4_comparison_across_thinking_and_sft_models.png}
%         % 四个模型在四个测试基准上的表现，其中Qwen2.5-VL-sft在我们的数据(MOSS-Video$_{train}$)上训练过
%         \caption{Performance of four models across four test benchmarks, with Qwen2.5-VL-sft trained on our dataset.}
%         \label{fig:zhuzhuang}   
%     \end{minipage}
% \end{figure}

\begin{wrapfigure}{r}{0.45\textwidth}
    \vspace{-7mm}
    \centering
    \includegraphics[width=0.45\textwidth]{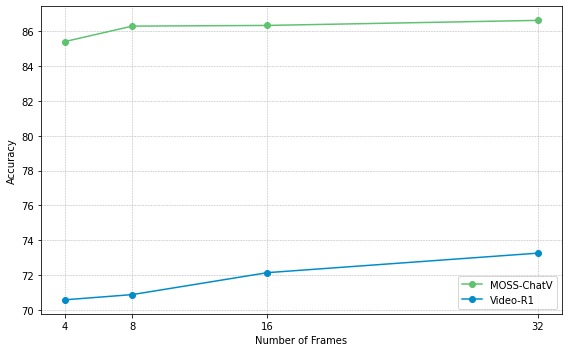}
    \caption{Performance impact of varying input frame counts.}
    \vspace{-2pt}
    \label{fig:frame_quality}
\end{wrapfigure}

%% file: figures/5_response_len.tex
\begin{figure*}
    \centering
    \includegraphics[width=1.0\linewidth]{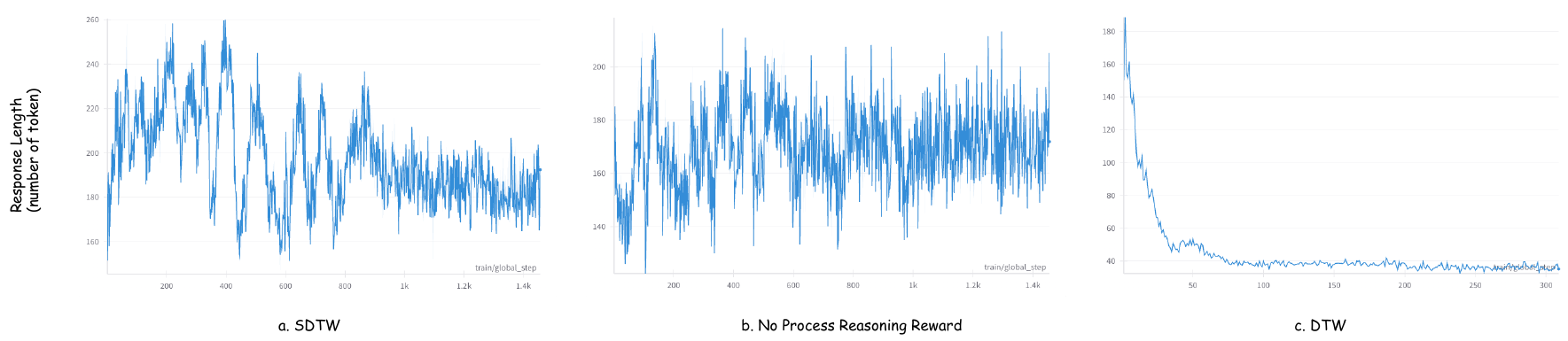}
    %（a）是用sdtw的训练过程中的响应长度，可以发现在训练初期，模型进行了广泛的尝试与探索，最终收敛到了一个合理的长度上（b）是在不使用推理过程监督的响应长度（c）是利用dtw的训练过程中的响应长度，可以发现出现了奖励劫持，从而大大降低了推理的长度
    \caption{Figure (a) shows that with Subsequence DTW (SDTW), response lengths initially fluctuate due to exploration but gradually converge to a stable range. Figure (b) reports training without process supervision, where response lengths remain unstable. Figure (c) illustrates Naive DTW, which induces reward hacking: the model shortens its reasoning drastically to exploit the distance metric.
}
    \label{fig:5_response_len}
\end{figure*}

%% file: tables/4_ablation.tex
\begin{table}[htbp]
\centering
\caption{Ablation Results}
\label{table_ablation}
\resizebox{0.8\textwidth}{!}{%
\begin{tabular}{lccc}
\toprule
\multicolumn{1}{c}{\textbf{Model}} & \textbf{MVBench} & \textbf{VideoMME} & \textbf{MOSS-Video} \\
\midrule
Qwen2.5-VL-7B                          & \uline{67.09}            & \uline{59.67}             & 67.00               \\
Qwen2.5-VL-7B+SFT (MOSS-ChatV-SFT)                                & 65.12 \(\textcolor{blue}{\downarrow _{\small\textbf{1.97}}}\)  & 55.24 \(\textcolor{blue}{\downarrow _{\small\textbf{4.43}}}\)  & 71.44 \(\textcolor{red}{\uparrow _{\small\textbf{4.44}}}\)   \\
Qwen2.5-VL-7B+T-GRPO (Video-R1)                    & 63.90 \(\textcolor{blue}{\downarrow _{\small\textbf{3.19}}}\)  & 59.30 \(\textcolor{blue}{\downarrow _{\small\textbf{0.37}}}\)  & 73.26 \(\textcolor{red}{\uparrow _{\small\textbf{6.26}}}\)   \\
Qwen2.5-VL-7B+GRPO (MOSS-ChatV-no-PPR)                    & 65.23 \(\textcolor{blue}{\downarrow _{\small\textbf{1.86}}}\)  & 55.30 \(\textcolor{blue}{\downarrow _{\small\textbf{4.37}}}\)  & \uline{84.17} \(\textcolor{red}{\uparrow _{\small\textbf{17.17}}}\)   \\
\rowcolor{myblue}
Qwen2.5-VL-7B+GRPO+Process Reasoning Reward (MOSS-ChatV)   & \textbf{67.60} \(\textcolor{red}{\uparrow _{\small\textbf{0.51}}}\)  & \textbf{59.96} \(\textcolor{red}{\uparrow _{\small\textbf{0.29}}}\)  & \textbf{86.62} \(\textcolor{red}{\uparrow _{\small\textbf{19.62}}}\)  \\
\bottomrule

\end{tabular}%
}
\end{table}

% \begin{table}[htbp]
% \centering
% \caption{Ablation Results}
% \label{table_ablation}
% \resizebox{1.0\textwidth}{!}{%
% \begin{tabular}{lccc}
% \toprule
% \multicolumn{1}{c}{\textbf{Model}} & \textbf{MVBench} & \textbf{VideoMME} & \textbf{MOSS-Video} \\
% \midrule
% Qwen2.5-VL-7B                          & \uline{67.09}            & \uline{59.67}             & 67.00               \\
% +SFT (MOSS-ChatV-SFT)                                & 65.12 \($\textcolor{blue}{\downarrow _{\small\textbf{1.97}}}$\)  & 55.24 \(\textcolor{blue}{\downarrow \small\textbf{4.43}}\)  & 71.44 \(\textcolor{red}{\uparrow \small\textbf{4.44}}\)   \\
% +T-GRPO (Video-R1)                    & 63.90 \(\textcolor{blue}{\downarrow \small\textbf{3.19}}\)  & 59.30 \(\textcolor{blue}{\downarrow \small\textbf{0.37}}\)  & 73.26 \(\textcolor{red}{\uparrow \small\textbf{6.26}}\)   \\
% +GRPO (MOSS-ChatV-no-PPR)                    & 65.23 \(\textcolor{blue}{\downarrow \small\textbf{1.86}}\)  & 55.30 \(\textcolor{blue}{\downarrow \small\textbf{4.37}}\)  & \uline{84.17} \(\textcolor{red}{\uparrow \small\textbf{17.17}}\)   \\
% \rowcolor{myblue}
% +GRPO+Process Reasoning Reward (MOSS-ChatV)   & \textbf{67.60} \(\textcolor{red}{\uparrow \small\textbf{0.51}}\)  & \textbf{59.96} \(\textcolor{red}{\uparrow \small\textbf{0.29}}\)  & \textbf{86.62} \(\textcolor{red}{\uparrow \small\textbf{19.62}}\)  \\
% \bottomrule

% \end{tabular}%
% }
% \end{table}

%% file: tables/4_MLLM_as_a_Judge.tex
\begin{table}[htbp]
\centering
\small
\setlength{\abovecaptionskip}{0.2cm} %调整caption与图的距离
\caption{MLLM as a judge for evaluating the performance of reasoning across different models.}
\label{MLLM as a judge}
\resizebox{0.8\textwidth}{!}{%
\begin{tabular}{lcccc}
\toprule
\textbf{Method} & \textbf{\makecell{Reasoning-Answer\\Consistency}} & \textbf{\makecell{Reasoning Content\\Repetitiveness}} & \textbf{\makecell{Logical Coherence \\\& Knowledge}} & \textbf{\makecell{Relevance to\\Video Content}} \\
\midrule

\textsc{Qwen2.5-VL}       & 0.69 & \textbf{8.87} & 6.97 & 6.82 \\
\textsc{Video-R1}        & \underline{0.78} & 4.14 & 6.87 & 6.57 \\
\rowcolor{myblue}
\textsc{MOSS-ChatV-no-PRR}   & 0.72 & \underline{7.80} & \textbf{7.80} & \textbf{7.45} \\
\rowcolor{myblue}
\textsc{MOSS-ChatV}            & \textbf{0.79} & 7.23 & \underline{7.59} & \underline{7.35} \\

\bottomrule
\end{tabular}
}
\end{table}

% \begin{table}[htbp]
% \centering
% \caption{Ablation Results}
% \label{table_ablation}
% \resizebox{1.0\textwidth}{!}{%
% \begin{tabular}{lccc}
% \toprule
% \multicolumn{1}{c}{\textbf{Model}} & \textbf{MVBench} & \textbf{VideoMME} & \textbf{MOSS-Video} \\
% \midrule
% Qwen2.5-VL-7B                          & \uline{67.09}            & \uline{59.67}             & 67.00               \\
% +SFT (MOSS-ChatV-SFT)                                & 65.12 \(\textcolor{blue}{\downarrow _{\small\textbf{1.97}}}\)  & 55.24 \(\textcolor{blue}{\downarrow _{\small\textbf{4.43}}}\)  & 71.44 \(\textcolor{red}{\uparrow _{\small\textbf{4.44}}}\)   \\
% +T-GRPO (Video-R1)                    & 63.90 \(\textcolor{blue}{\downarrow _{\small\textbf{3.19}}}\)  & 59.30 \(\textcolor{blue}{\downarrow _{\small\textbf{0.37}}}\)  & 73.26 \(\textcolor{red}{\uparrow _{\small\textbf{6.26}}}\)   \\
% +GRPO (MOSS-ChatV-no-PPR)                    & 65.23 \(\textcolor{blue}{\downarrow _{\small\textbf{1.86}}}\)  & 55.30 \(\textcolor{blue}{\downarrow _{\small\textbf{4.37}}}\)  & \uline{84.17} \(\textcolor{red}{\uparrow _{\small\textbf{17.17}}}\)   \\
% \rowcolor{myblue}
% +GRPO+Process Reasoning Reward (MOSS-ChatV)   & \textbf{67.60} \(\textcolor{red}{\uparrow _{\small\textbf{0.51}}}\)  & \textbf{59.96} \(\textcolor{red}{\uparrow _{\small\textbf{0.29}}}\)  & \textbf{86.62} \(\textcolor{red}{\uparrow _{\small\textbf{19.62}}}\)  \\
% \bottomrule

% \end{tabular}%
% }
% \end{table}

%% file: sections/5_related_work.tex
\section{Related Work} 
\subsection{Advanced Video-LLM}
With the burgeoning development of Multimodal Large Language Models (MLLMs), such as Qwen~\citep{wang2024qwen2} and InternVL~\citep{wang2024internvideo2, wang2025internvideo2}, video understanding has emerged as a critical dimension for evaluating model capabilities. To enhance models' comprehension of video content, researchers have employed a variety of strategies. For instance, VideoChatGPT~\citep{maaz2023video} focuses on improving model proficiency in video dialogue, description, and reasoning by introducing video-specific instruction-tuning datasets and a quantitative evaluation framework. Other approaches, exemplified by models like NVILA~\citep{liu2024nvila}, LongVU~\citep{shen2024longvu}, and VideoLLaMA3~\citep{zhang2025videollama}, enhance their capacity to process long videos through various visual token compression techniques, such as removing redundant tokens or employing MLP-based compression. Furthermore, models such as LLaVA-OV~\citep{li2024llava} are typically pre-trained on large-scale video-text pair datasets (video training data) and subsequently fine-tuned using instruction data for tasks like video question answering and description generation to adapt to diverse video understanding scenarios. These works collectively provide an excellent foundation for advancing video reasoning capabilities in Video-LLMs. While these methods advance general video understanding, our work introduces a reinforcement learning framework to directly supervise the temporal reasoning process.

\subsection{Reasoning and Reinforcement Learning in Video-LLMs}
To enhance the reasoning capabilities of video models, researchers have made numerous attempts, such as utilizing rationale construction, structural reasoning, objective granularity, and other methods~\citep{wang2025multimodalchainofthoughtreasoningcomprehensive}.
Recent advances in Reinforcement Learning (RL) have significantly improved LLM alignment and specialized capabilities, as seen in reasoning LLMs~\citep{deepseekr1}. This success has spurred RL-based enhancements for Multimodal LLMs~\citep{yang2025r1onevisionadvancinggeneralizedmultimodal, meng2025mmeurekaexploringfrontiersmultimodal}. Specifically, for video modality, Videochat- R1~\citep{li2025videochat} and TimeZero~\citep{wang2025timezero} leverage RL rewards for temporal grounding, while TinyLLaVA-Video-R1~\citep{zhang2025tinyllava} demonstrates RL’s effectiveness even on small models. Video-R1~\citep{videor1} employs contrastive RL to improve temporal understanding. Our approach is distinguished by a novel, rule-based Process Reasoning Reward (PRR), which offers more granular supervision on the reasoning path itself.

%% file: sections/6_conclusion.tex
\section{Conclusion}
%我们通过分析视频状态预测任务和视频推理能力的关系，认为两者有相互促进作用。基于这个想法，我们提供了对应的视频状态预测数据集MOSS-Video用于训练和测试。此外，针对视频模态的强化微调，我们提出了基于规则的Process Reassing Reward。我们通过对比和消融实验，证明了我们的数据和方法的有效性，在单一任务训练数据的情况下做到了整体视频分析能力的提升，并且在推理内容质量上也体现了稳定。综上所诉，我们发现在视频语境下的模型推理质量应该受到重视，通过MOSS-ChatV验证了强化微调过程监督对视频推理能力的提升有显著帮助,并在低质量视频输入的语境下做到了涨点和SOTA。
Through analyzing the relationship between video state prediction tasks and video reasoning capabilities, we demonstrate their mutual reinforcement. Based on this insight, we introduce MOSS-Video, a dedicated dataset for training and evaluating video state prediction task. For reinforcement fine-tuning of video modalities, we propose Process Reasoning Reward (PRR), a rule-based reward mechanism. Comparative and ablation experiments confirm the effectiveness of our approach. Using single-task training data alone, we achieve holistic improvements in video analysis performance while maintaining stable reasoning quality. In summary, we find that model reasoning capability in video contexts deserves greater attention. Through MOSS-ChatV, we verify that reinforcement fine-tuning with process supervision significantly enhances video reasoning performance, achieving performance gains and state-of-the-art results even under low-quality video inputs.

\newpage
\section{Ethics Statement}
This research complies with ethical standards. It utilizes datasets that are either synthetic or publicly available, and contains no sensitive or personally identifiable information. The study involves no direct human subjects, nor does it pose any privacy or security concerns. All methodologies and experiments were conducted in accordance with applicable laws and established research integrity practices.
There are no conflicts of interest, no undue influence from external sponsorship, and no concerns related to discrimination, bias, or fairness. Moreover, this research does not lead to any harmful insights or applications.

\section{Reproducibility Statement}
We have taken steps to ensure the reproducibility of the results presented in this paper. The experimental settings, including datasets and model designs, are thoroughly described in Section~\ref{sec:experiment}. Source code will be made publicly available upon acceptance.

\section{LLM Usage Statement}
In this work, large language models (LLMs) were used exclusively to assist with writing, editing, and LaTeX formatting. Their role was confined to enhancing clarity, grammar, and overall presentation; they had no impact on the design of experiments, data processing, analysis, or the interpretation of results.

% \section{Limitations and Future Works}
% %未来我们可以扩大MOSS-Video的规模，使其囊括更多的任务。当前策略还是需要借助标注数据的推理过程，未来可以继续探索强化微调过程监督的策略，从而减少对外部过程监督信号的需求
% %我们认为MLLM as a Judge应该有更多的开发潜力，现目前我们只能做到prompt-based的质量检测方法，未来可以继续探索更多的针对性的裁判模型，从而提升推理内容质量检测的质量。
% In the future, we can scale up MOSS-Video to encompass more tasks. The current process supervision strategy still relies on labeled data for the reasoning process, but we can further explore reinforcement fine-tuning with process supervision to reduce dependence on external supervision signals.

% We believe that MLLM as a Judge holds greater potential for development in the video scenarios. Currently, we only implement prompt-based quality evaluation methods, but in the future, we can explore more specialized judge models to improve the quality assessment of reasoning outputs.

%% file: sections/appendix.tex
\counterwithin{figure}{section}  
\counterwithin{table}{section}   
\renewcommand{\thefigure}{\thesection.\arabic{figure}}
\renewcommand{\thetable}{\thesection.\arabic{table}}

% The appendix includes the following sections:
% \begin{itemize}
%     \item \textbf{Section~\ref{appendix:experimental Analysis Supplement}: Experimental Analysis Supplement.}
%     \item \textbf{Section~\ref{appendix:Framework Design and Application Extensions}:Framework Design and Application Extensions.}
%     \item \textbf{Section~\ref{appendix:case-study}: Case.}
% \end{itemize}

\section{Details of MOSS-Video}
\label{sec:appendix_details_of_moss_video}
%% 我们选择高质量的ShareGPT4Video作为数据来源。我们有两个并行的数据管道从ShareGPT4Video分别得到粗粒度和细粒度的标注信息，具体而言，我们使用GPT4-o对视频的标注文件进行信息提取，在粗粒度的数据管道中我们可以得到<Object, State, Timestamp>三元组，表示某个Object在某个时间段内的State；在细粒度的数据管道中我们可以得到<Object, State:Description, Timestamp>三元组，表示某个Object在某个时间段内处于某种State时的具体描述。在得到两个三元组信息后，我们再次使用GPT4-o，使其根据Object在各个时刻的状态进行建模，最终得到QA对。QA对中的Question是对物体未来时刻的查询。
We leverage high-quality ShareGPT4Video as our primary data source, employing two parallel annotation pipelines to capture both coarse- and fine-grained object states. In the coarse-grained pipeline, GPT4-o processes each video’s annotation file to produce triplets of the form ⟨Object, State, Timestamp⟩, thereby characterizing an object’s state over a defined temporal interval. Concurrently, in the fine-grained pipeline, GPT4-o extracts more detailed triplets ⟨Object, State: Description, Timestamp⟩, which enrich each state with a specific textual description at a precise moment. Finally, we again invoke GPT4-o to integrate these two annotation streams into a unified temporal model of object dynamics, from which we automatically generate question–answer pairs that probe the predicted future states of objects.
\begin{figure}[htbp]
    \centering
    \includegraphics[width=1\linewidth]{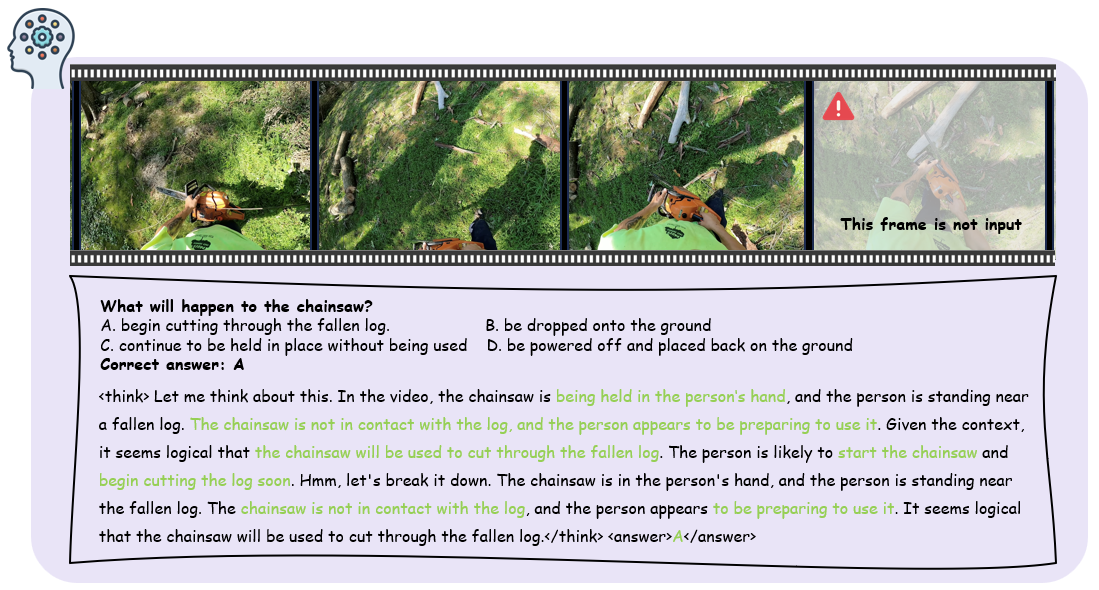}
    \caption{The example of MOSS-Video.}
    \label{fig:enter-label}
\end{figure}

\begin{figure*}[htbp]
\centering
\begin{tcolorbox}[
    colback=myblue!5!white,
    colframe=black!50!myblue,
    coltitle=black,
    colbacktitle=myblue!20!white,
    fonttitle=\bfseries,
    title=Prompts for Reasoning Process Evaluation,
]
Given the following video captions, extract object-centric information for a video prediction task.

\medskip
\textbf{Reference Video Segments (what can be seen):}
\begin{verbatim}
for caption in reference_captions:
    prompt += f"Time {caption['time_stamp']}s:
    {caption['content']}\n\n"
\end{verbatim}

\medskip
\textbf{Prediction Segment (what needs to be predicted):}
\begin{verbatim}
Time {prediction_caption['time_stamp']}s: 
{prediction_caption['content']}
\end{verbatim}

\medskip
\textbf{Task 1:} For each object in the video, provide coarse-grained information as \textless object, state, time\textgreater{} triplets for both reference and prediction segments.

\smallskip
\textbf{Task 2:} For each object, provide fine-grained information as \textless object, change description, time\textgreater{} triplets for how objects change over time.

\medskip
\textbf{Format your response as follows:}

\emph{COARSE-GRAINED INFORMATION:}

Object 1:
\begin{itemize}
  \item At time [timestamp]: [state description]
  \item At time [timestamp]: [state description]
  \item \dots
\end{itemize}

Object 2:
\begin{itemize}
  \item At time [timestamp]: [state description]
  \item \dots
\end{itemize}

\medskip
\emph{FINE-GRAINED INFORMATION:}

Object 1:
\begin{itemize}
  \item From time [start] to [end]: [detailed change description]
  \item \dots
\end{itemize}

Object 2:
\begin{itemize}
  \item From time [start] to [end]: [detailed change description]
  \item \dots
\end{itemize}

Focus only on objects that appear in the prediction segment. Be specific and detailed in your descriptions.
\end{tcolorbox}
\caption{Prompt template used for evaluating the reasoning process of video question answering models.}
\label{fig:prompt_for_evaluating_reasoning}
\end{figure*}

\section{MLLM-as-a-Judge for Reasponse Quality}
%% 我们使用了GPT4-o对推理模型的推理过程和最终响应结果进行了核验，具体使用的prompt可参考~\ref{fig:prompt_for_evaluating_reasoning}
We used GPT-4-o to verify the reasoning process and the final response results of the reasoning model. The specific evaluated dimensions are listed below:

\textbf{Reasoning-Answer Consistency (0 or 1)}: This is a binary metric. A score of 1 is awarded if the final conclusion of the reasoning aligns with the content of the model-selected option; otherwise, it receives a score of 0.

\textbf{Reasoning Content Repetitiveness (0-10)}: This assesses the presence of redundant information in the reasoning process. Higher repetitiveness results in a lower score, aiming to measure information density and avoid the amplification of potential biases or errors.

\textbf{Logical Coherence and Knowledge Accuracy (0-10)}: This directly evaluates the intrinsic quality of the reasoning process. The more rigorous the logic and the more accurate the application of world knowledge, the higher the score.

\textbf{Reasoning-Video Content Relevance (0-10)}: This measures how closely the reasoning is based on the video content. Higher relevance yields a higher score, aiming to penalize unfounded speculations or associations unrelated to the video.

In addition, the specificprompts used can refer to ~\ref{fig:prompt_for_evaluating_reasoning}.
\input{figures/supp_prompt_eval}
%% 

%% file: figures/supp_prompt_eval.tex
\begin{figure*}[htbp]
\centering
\begin{tcolorbox}[
    colback=myblue!5!white,
    colframe=black!50!myblue,
    coltitle=black,
    colbacktitle=myblue!20!white,
    fonttitle=\bfseries,
    title=Prompts for Reasoning Process Evaluation,
]
\small
You are a professional video question answering reasoning process evaluator. Your task is to evaluate the quality of the reasoning process \textbf{ONLY}, based on the provided video frames, question, options, and a model's reasoning text. You \textbf{DO NOT} need to judge the correctness of the model's final answer.\\[1mm]

Please evaluate the reasoning process based on the following dimensions:\\[1mm]

\textbf{1. Reasoning Conclusion and Answer Tag Consistency (0 or 1 point):}\\
Criterion: Check whether the conclusion in the reasoning text semantically matches the option marked by the \texttt{<answer>} tag. You should carefully analyze and consider the logic within the \texttt{<think>} tag.\\
Scoring:\\
\quad 1 point: Consistent.\\
\quad 0 points: Inconsistent.\\[1mm]

\textbf{2. Reasoning Content Repetitiveness (0–10 points, lower score for more repetition):}\\
Criterion: Assess whether the reasoning content contains unnecessary repetition of words, phrases, or semantics.\\
Scoring Guidelines:\\
\quad 9–10 points: Very concise, no unnecessary repetition, high information density.\\
\quad 6–8 points: Slight repetition or reasonable restatement for emphasis, overall flow is smooth.\\
\quad 3–5 points: Obvious repetition, but core idea is still discernible.\\
\quad 0–2 points: Massive repetition, almost no new information.\\[1mm]

\textbf{3. Reasoning Logical Coherence and Knowledge Accuracy (0–10 points):}\\
Criterion: Evaluate if the reasoning steps are clear and coherent, the logical chain complete, and any assumptions reasonable and correct.\\
Scoring Guidelines:\\
\quad 9–10 points: Rigorous logic, well-organized, sufficient argumentation, accurate assumptions.\\
\quad 6–8 points: Generally coherent with minor flaws.\\
\quad 3–5 points: Obvious breaks or minor errors not affecting main conclusion.\\
\quad 0–2 points: Chaotic or contradictory logic, erroneous assumptions.\\[1mm]

\textbf{4. Reasoning and Video Content Relevance (0–10 points, lower score for more deviation):}\\
Criterion: Assess whether observations and conclusions are strictly based on provided video frames.\\
Scoring Guidelines:\\
\quad 9–10 points: Strictly based on video content with strong evidence.\\
\quad 6–8 points: Primarily based on content with minor reasonable inference.\\
\quad 3–5 points: Mostly imagination or misunderstanding of video.\\
\quad 0–2 points: Completely unrelated or speculative.\\[1mm]

\textbf{[Input Information]}\\[1mm]
Video Frames: \{num\_frames\_provided\} frames are provided. (Actual frames are sent as image data)\\
Question: \{question\_text\}\\
Model Reasoning Text: \{model\_reasoning\_text\}\\
Model's Answer Tag Content: \texttt{<answer>\{model\_answer\_tag\_content\}</answer>}\\[1mm]

\textbf{[Your Evaluation Output]}\\[1mm]
Please provide your evaluation scores strictly in the following format, one line per dimension, containing only the score:\\[1mm]
\texttt{Dimension1\_Score: [0 or 1]}\\
\texttt{Dimension2\_Score: [0–10]}\\
\texttt{Dimension3\_Score: [0–10]}\\
\texttt{Dimension4\_Score: [0–10]}
\end{tcolorbox}
\caption{Prompt template used for evaluating the reasoning process of video question answering models.}
\label{fig:prompt_for_evaluating_reasoning}
\end{figure*}

%% file: main.bbl
\begin{thebibliography}{32}
\providecommand{\natexlab}[1]{#1}
\providecommand{\url}[1]{\texttt{#1}}
\expandafter\ifx\csname urlstyle\endcsname\relax
  \providecommand{\doi}[1]{doi: #1}\else
  \providecommand{\doi}{doi: \begingroup \urlstyle{rm}\Url}\fi

\bibitem[Bai et~al.(2025)Bai, Chen, Liu, Wang, Ge, Song, Dang, Wang, Wang, Tang, Zhong, Zhu, Yang, Li, Wan, Wang, Ding, Fu, Xu, Ye, Zhang, Xie, Cheng, Zhang, Yang, Xu, and Lin]{bai2025qwen25vltechnicalreport}
Shuai Bai, Keqin Chen, Xuejing Liu, Jialin Wang, Wenbin Ge, Sibo Song, Kai Dang, Peng Wang, Shijie Wang, Jun Tang, Humen Zhong, Yuanzhi Zhu, Mingkun Yang, Zhaohai Li, Jianqiang Wan, Pengfei Wang, Wei Ding, Zheren Fu, Yiheng Xu, Jiabo Ye, Xi~Zhang, Tianbao Xie, Zesen Cheng, Hang Zhang, Zhibo Yang, Haiyang Xu, and Junyang Lin.
\newblock Qwen2.5-vl technical report, 2025.
\newblock URL \url{https://arxiv.org/abs/2502.13923}.

\bibitem[Caffagni et~al.(2024)Caffagni, Cocchi, Barsellotti, Moratelli, Sarto, Baraldi, Cornia, and Cucchiara]{caffagni2024revolution}
Davide Caffagni, Federico Cocchi, Luca Barsellotti, Nicholas Moratelli, Sara Sarto, Lorenzo Baraldi, Marcella Cornia, and Rita Cucchiara.
\newblock The revolution of multimodal large language models: a survey.
\newblock \emph{arXiv preprint arXiv:2402.12451}, 2024.

\bibitem[Chen et~al.(2024)Chen, Wei, Li, Dong, Zhang, Zang, Chen, Duan, Lin, Tang, Yuan, Qiao, Lin, Zhao, and Wang]{chen2024sharegpt4video}
Lin Chen, Xilin Wei, Jinsong Li, Xiaoyi Dong, Pan Zhang, Yuhang Zang, Zehui Chen, Haodong Duan, Bin Lin, Zhenyu Tang, Li~Yuan, Yu~Qiao, Dahua Lin, Feng Zhao, and Jiaqi Wang.
\newblock Sharegpt4video: Improving video understanding and generation with better captions.
\newblock \emph{arXiv preprint arXiv:2406.04325}, 2024.

\bibitem[Cheng et~al.(2024)Cheng, Leng, Zhang, Xin, Li, Chen, Zhu, Zhang, Luo, Zhao, et~al.]{cheng2024videollama}
Zesen Cheng, Sicong Leng, Hang Zhang, Yifei Xin, Xin Li, Guanzheng Chen, Yongxin Zhu, Wenqi Zhang, Ziyang Luo, Deli Zhao, et~al.
\newblock Videollama 2: Advancing spatial-temporal modeling and audio understanding in video-llms.
\newblock \emph{CoRR}, 2024.

\bibitem[DeepSeek-AI et~al.(2025)DeepSeek-AI, Guo, Yang, Zhang, Song, Zhang, Xu, Zhu, Ma, Wang, Bi, Zhang, Yu, Wu, Wu, Gou, Shao, Li, Gao, Liu, Xue, Wang, Wu, Feng, Lu, Zhao, Deng, Zhang, Ruan, Dai, Chen, Ji, Li, Lin, Dai, Luo, Hao, Chen, Li, Zhang, Bao, Xu, Wang, Ding, Xin, Gao, Qu, Li, Guo, Li, Wang, Chen, Yuan, Qiu, Li, Cai, Ni, Liang, Chen, Dong, Hu, Gao, Guan, Huang, Yu, Wang, Zhang, Zhao, Wang, Zhang, Xu, Xia, Zhang, Zhang, Tang, Li, Wang, Li, Tian, Huang, Zhang, Wang, Chen, Du, Ge, Zhang, Pan, Wang, Chen, Jin, Chen, Lu, Zhou, Chen, Ye, Wang, Yu, Zhou, Pan, Li, Zhou, Wu, Ye, Yun, Pei, Sun, Wang, Zeng, Zhao, Liu, Liang, Gao, Yu, Zhang, Xiao, An, Liu, Wang, Chen, Nie, Cheng, Liu, Xie, Liu, Yang, Li, Su, Lin, Li, Jin, Shen, Chen, Sun, Wang, Song, Zhou, Wang, Shan, Li, Wang, Wei, Zhang, Xu, Li, Zhao, Sun, Wang, Yu, Zhang, Shi, Xiong, He, Piao, Wang, Tan, Ma, Liu, Guo, Ou, Wang, Gong, Zou, He, Xiong, Luo, You, Liu, Zhou, Zhu, Xu, Huang, Li, Zheng, Zhu, Ma, Tang, Zha, Yan, Ren, Ren, Sha, Fu, Xu, Xie, Zhang,
  Hao, Ma, Yan, Wu, Gu, Zhu, Liu, Li, Xie, Song, Pan, Huang, Xu, Zhang, and Zhang]{deepseekr1}
DeepSeek-AI, Daya Guo, Dejian Yang, Haowei Zhang, Junxiao Song, Ruoyu Zhang, Runxin Xu, Qihao Zhu, Shirong Ma, Peiyi Wang, Xiao Bi, Xiaokang Zhang, Xingkai Yu, Yu~Wu, Z.~F. Wu, Zhibin Gou, Zhihong Shao, Zhuoshu Li, Ziyi Gao, Aixin Liu, Bing Xue, Bingxuan Wang, Bochao Wu, Bei Feng, Chengda Lu, Chenggang Zhao, Chengqi Deng, Chenyu Zhang, Chong Ruan, Damai Dai, Deli Chen, Dongjie Ji, Erhang Li, Fangyun Lin, Fucong Dai, Fuli Luo, Guangbo Hao, Guanting Chen, Guowei Li, H.~Zhang, Han Bao, Hanwei Xu, Haocheng Wang, Honghui Ding, Huajian Xin, Huazuo Gao, Hui Qu, Hui Li, Jianzhong Guo, Jiashi Li, Jiawei Wang, Jingchang Chen, Jingyang Yuan, Junjie Qiu, Junlong Li, J.~L. Cai, Jiaqi Ni, Jian Liang, Jin Chen, Kai Dong, Kai Hu, Kaige Gao, Kang Guan, Kexin Huang, Kuai Yu, Lean Wang, Lecong Zhang, Liang Zhao, Litong Wang, Liyue Zhang, Lei Xu, Leyi Xia, Mingchuan Zhang, Minghua Zhang, Minghui Tang, Meng Li, Miaojun Wang, Mingming Li, Ning Tian, Panpan Huang, Peng Zhang, Qiancheng Wang, Qinyu Chen, Qiushi Du, Ruiqi Ge, Ruisong
  Zhang, Ruizhe Pan, Runji Wang, R.~J. Chen, R.~L. Jin, Ruyi Chen, Shanghao Lu, Shangyan Zhou, Shanhuang Chen, Shengfeng Ye, Shiyu Wang, Shuiping Yu, Shunfeng Zhou, Shuting Pan, S.~S. Li, Shuang Zhou, Shaoqing Wu, Shengfeng Ye, Tao Yun, Tian Pei, Tianyu Sun, T.~Wang, Wangding Zeng, Wanjia Zhao, Wen Liu, Wenfeng Liang, Wenjun Gao, Wenqin Yu, Wentao Zhang, W.~L. Xiao, Wei An, Xiaodong Liu, Xiaohan Wang, Xiaokang Chen, Xiaotao Nie, Xin Cheng, Xin Liu, Xin Xie, Xingchao Liu, Xinyu Yang, Xinyuan Li, Xuecheng Su, Xuheng Lin, X.~Q. Li, Xiangyue Jin, Xiaojin Shen, Xiaosha Chen, Xiaowen Sun, Xiaoxiang Wang, Xinnan Song, Xinyi Zhou, Xianzu Wang, Xinxia Shan, Y.~K. Li, Y.~Q. Wang, Y.~X. Wei, Yang Zhang, Yanhong Xu, Yao Li, Yao Zhao, Yaofeng Sun, Yaohui Wang, Yi~Yu, Yichao Zhang, Yifan Shi, Yiliang Xiong, Ying He, Yishi Piao, Yisong Wang, Yixuan Tan, Yiyang Ma, Yiyuan Liu, Yongqiang Guo, Yuan Ou, Yuduan Wang, Yue Gong, Yuheng Zou, Yujia He, Yunfan Xiong, Yuxiang Luo, Yuxiang You, Yuxuan Liu, Yuyang Zhou, Y.~X. Zhu,
  Yanhong Xu, Yanping Huang, Yaohui Li, Yi~Zheng, Yuchen Zhu, Yunxian Ma, Ying Tang, Yukun Zha, Yuting Yan, Z.~Z. Ren, Zehui Ren, Zhangli Sha, Zhe Fu, Zhean Xu, Zhenda Xie, Zhengyan Zhang, Zhewen Hao, Zhicheng Ma, Zhigang Yan, Zhiyu Wu, Zihui Gu, Zijia Zhu, Zijun Liu, Zilin Li, Ziwei Xie, Ziyang Song, Zizheng Pan, Zhen Huang, Zhipeng Xu, Zhongyu Zhang, and Zhen Zhang.
\newblock Deepseek-r1: Incentivizing reasoning capability in llms via reinforcement learning, 2025.
\newblock URL \url{https://arxiv.org/abs/2501.12948}.

\bibitem[Fei et~al.(2024)Fei, Wu, Ji, Zhang, Zhang, Lee, and Hsu]{fei2024videoofthoughtstepbystepvideoreasoning}
Hao Fei, Shengqiong Wu, Wei Ji, Hanwang Zhang, Meishan Zhang, Mong-Li Lee, and Wynne Hsu.
\newblock Video-of-thought: Step-by-step video reasoning from perception to cognition, 2024.
\newblock URL \url{https://arxiv.org/abs/2501.03230}.

\bibitem[Feng et~al.(2025)Feng, Gong, Li, Guo, Wang, Peng, Wu, Zhang, Wang, and Yue]{videor1}
Kaituo Feng, Kaixiong Gong, Bohao Li, Zonghao Guo, Yibing Wang, Tianshuo Peng, Junfei Wu, Xiaoying Zhang, Benyou Wang, and Xiangyu Yue.
\newblock Video-r1: Reinforcing video reasoning in mllms, 2025.
\newblock URL \url{https://arxiv.org/abs/2503.21776}.

\bibitem[Fu et~al.(2024)Fu, Dai, Luo, Li, Ren, Zhang, Wang, Zhou, Shen, Zhang, et~al.]{fu2024video}
Chaoyou Fu, Yuhan Dai, Yongdong Luo, Lei Li, Shuhuai Ren, Renrui Zhang, Zihan Wang, Chenyu Zhou, Yunhang Shen, Mengdan Zhang, et~al.
\newblock Video-mme: The first-ever comprehensive evaluation benchmark of multi-modal llms in video analysis.
\newblock \emph{arXiv preprint arXiv:2405.21075}, 2024.

\bibitem[Li et~al.(2024{\natexlab{a}})Li, Zhang, Guo, Zhang, Li, Zhang, Zhang, Zhang, Li, Liu, et~al.]{li2024llava}
Bo~Li, Yuanhan Zhang, Dong Guo, Renrui Zhang, Feng Li, Hao Zhang, Kaichen Zhang, Peiyuan Zhang, Yanwei Li, Ziwei Liu, et~al.
\newblock Llava-onevision: Easy visual task transfer.
\newblock \emph{arXiv preprint arXiv:2408.03326}, 2024{\natexlab{a}}.

\bibitem[Li et~al.(2024{\natexlab{b}})Li, Wang, He, Li, Wang, Liu, Wang, Xu, Chen, Luo, et~al.]{li2024mvbench}
Kunchang Li, Yali Wang, Yinan He, Yizhuo Li, Yi~Wang, Yi~Liu, Zun Wang, Jilan Xu, Guo Chen, Ping Luo, et~al.
\newblock Mvbench: A comprehensive multi-modal video understanding benchmark.
\newblock In \emph{Proceedings of the IEEE/CVF Conference on Computer Vision and Pattern Recognition}, pp.\  22195--22206, 2024{\natexlab{b}}.

\bibitem[Li et~al.(2025)Li, Yan, Meng, Dong, Zeng, He, Wang, Qiao, Wang, and Wang]{li2025videochat}
Xinhao Li, Ziang Yan, Desen Meng, Lu~Dong, Xiangyu Zeng, Yinan He, Yali Wang, Yu~Qiao, Yi~Wang, and Limin Wang.
\newblock Videochat-r1: Enhancing spatio-temporal perception via reinforcement fine-tuning.
\newblock \emph{arXiv preprint arXiv:2504.06958}, 2025.

\bibitem[Liang et~al.(2024)Liang, Xu, Hong, Shang, Wang, Fu, and Liu]{liang2024survey}
Zijing Liang, Yanjie Xu, Yifan Hong, Penghui Shang, Qi~Wang, Qiang Fu, and Ke~Liu.
\newblock A survey of multimodel large language models.
\newblock In \emph{Proceedings of the 3rd International Conference on Computer, Artificial Intelligence and Control Engineering}, pp.\  405--409, 2024.

\bibitem[Liu et~al.(2024{\natexlab{a}})Liu, Li, Liu, Wang, Ren, Li, Chen, Sun, and Hou]{liu2024tempcompassvideollmsreally}
Yuanxin Liu, Shicheng Li, Yi~Liu, Yuxiang Wang, Shuhuai Ren, Lei Li, Sishuo Chen, Xu~Sun, and Lu~Hou.
\newblock Tempcompass: Do video llms really understand videos?, 2024{\natexlab{a}}.
\newblock URL \url{https://arxiv.org/abs/2403.00476}.

\bibitem[Liu et~al.(2024{\natexlab{b}})Liu, Zhu, Shi, Zhang, Lou, Yang, Xi, Cao, Gu, Li, et~al.]{liu2024nvila}
Zhijian Liu, Ligeng Zhu, Baifeng Shi, Zhuoyang Zhang, Yuming Lou, Shang Yang, Haocheng Xi, Shiyi Cao, Yuxian Gu, Dacheng Li, et~al.
\newblock Nvila: Efficient frontier visual language models.
\newblock \emph{arXiv preprint arXiv:2412.04468}, 2024{\natexlab{b}}.

\bibitem[Maaz et~al.(2023)Maaz, Rasheed, Khan, and Khan]{maaz2023video}
Muhammad Maaz, Hanoona Rasheed, Salman Khan, and Fahad~Shahbaz Khan.
\newblock Video-chatgpt: Towards detailed video understanding via large vision and language models.
\newblock \emph{arXiv preprint arXiv:2306.05424}, 2023.

\bibitem[Meng et~al.(2025)Meng, Du, Liu, Zhou, Lu, Fu, Han, Shi, Wang, He, Zhang, Luo, Qiao, Zhang, and Shao]{meng2025mmeurekaexploringfrontiersmultimodal}
Fanqing Meng, Lingxiao Du, Zongkai Liu, Zhixiang Zhou, Quanfeng Lu, Daocheng Fu, Tiancheng Han, Botian Shi, Wenhai Wang, Junjun He, Kaipeng Zhang, Ping Luo, Yu~Qiao, Qiaosheng Zhang, and Wenqi Shao.
\newblock Mm-eureka: Exploring the frontiers of multimodal reasoning with rule-based reinforcement learning, 2025.
\newblock URL \url{https://arxiv.org/abs/2503.07365}.

\bibitem[Shao et~al.(2024)Shao, Wang, Zhu, Xu, Song, Bi, Zhang, Zhang, Li, Wu, and Guo]{shao2024deepseekmathpushinglimitsmathematical}
Zhihong Shao, Peiyi Wang, Qihao Zhu, Runxin Xu, Junxiao Song, Xiao Bi, Haowei Zhang, Mingchuan Zhang, Y.~K. Li, Y.~Wu, and Daya Guo.
\newblock Deepseekmath: Pushing the limits of mathematical reasoning in open language models, 2024.
\newblock URL \url{https://arxiv.org/abs/2402.03300}.

\bibitem[Shen et~al.(2024)Shen, Xiong, Zhao, Wu, Chen, Zhu, Liu, Xiao, Varadarajan, Bordes, et~al.]{shen2024longvu}
Xiaoqian Shen, Yunyang Xiong, Changsheng Zhao, Lemeng Wu, Jun Chen, Chenchen Zhu, Zechun Liu, Fanyi Xiao, Balakrishnan Varadarajan, Florian Bordes, et~al.
\newblock Longvu: Spatiotemporal adaptive compression for long video-language understanding.
\newblock \emph{arXiv preprint arXiv:2410.17434}, 2024.

\bibitem[Wang et~al.(2024{\natexlab{a}})Wang, Bai, Tan, Wang, Fan, Bai, Chen, Liu, Wang, Ge, et~al.]{wang2024qwen2}
Peng Wang, Shuai Bai, Sinan Tan, Shijie Wang, Zhihao Fan, Jinze Bai, Keqin Chen, Xuejing Liu, Jialin Wang, Wenbin Ge, et~al.
\newblock Qwen2-vl: Enhancing vision-language model's perception of the world at any resolution.
\newblock \emph{arXiv preprint arXiv:2409.12191}, 2024{\natexlab{a}}.

\bibitem[Wang \& Peng(2025)Wang and Peng]{wang-2025-open-r1-video}
Xiaodong Wang and Peixi Peng.
\newblock Open-r1-video, 2025.

\bibitem[Wang et~al.(2025{\natexlab{a}})Wang, Wu, Zhang, Yan, Liu, Luo, and Fei]{wang2025multimodalchainofthoughtreasoningcomprehensive}
Yaoting Wang, Shengqiong Wu, Yuecheng Zhang, Shuicheng Yan, Ziwei Liu, Jiebo Luo, and Hao Fei.
\newblock Multimodal chain-of-thought reasoning: A comprehensive survey, 2025{\natexlab{a}}.
\newblock URL \url{https://arxiv.org/abs/2503.12605}.

\bibitem[Wang et~al.(2025{\natexlab{b}})Wang, Xu, Yue, Xiao, Wang, Zhang, Yang, Wang, and Jin]{wang2025timezero}
Ye~Wang, Boshen Xu, Zihao Yue, Zihan Xiao, Ziheng Wang, Liang Zhang, Dingyi Yang, Wenxuan Wang, and Qin Jin.
\newblock Timezero: Temporal video grounding with reasoning-guided lvlm.
\newblock \emph{arXiv preprint arXiv:2503.13377}, 2025{\natexlab{b}}.

\bibitem[Wang et~al.(2024{\natexlab{b}})Wang, Li, Li, Yu, He, Chen, Pei, Zheng, Wang, Shi, et~al.]{wang2024internvideo2}
Yi~Wang, Kunchang Li, Xinhao Li, Jiashuo Yu, Yinan He, Guo Chen, Baoqi Pei, Rongkun Zheng, Zun Wang, Yansong Shi, et~al.
\newblock Internvideo2: Scaling foundation models for multimodal video understanding.
\newblock In \emph{European Conference on Computer Vision}, pp.\  396--416. Springer, 2024{\natexlab{b}}.

\bibitem[Wang et~al.(2025{\natexlab{c}})Wang, Li, Yan, He, Yu, Zeng, Wang, Ma, Huang, Gao, et~al.]{wang2025internvideo2}
Yi~Wang, Xinhao Li, Ziang Yan, Yinan He, Jiashuo Yu, Xiangyu Zeng, Chenting Wang, Changlian Ma, Haian Huang, Jianfei Gao, et~al.
\newblock Internvideo2. 5: Empowering video mllms with long and rich context modeling.
\newblock \emph{arXiv preprint arXiv:2501.12386}, 2025{\natexlab{c}}.

\bibitem[Xiao et~al.(2021)Xiao, Shang, Yao, and Chua]{nextqa}
Junbin Xiao, Xindi Shang, Angela Yao, and Tat-Seng Chua.
\newblock Next-qa: Next phase of question-answering to explaining temporal actions.
\newblock In \emph{Proceedings of the IEEE/CVF conference on computer vision and pattern recognition}, pp.\  9777--9786, 2021.

\bibitem[Xun et~al.(2025)Xun, Tao, Li, Shi, Lin, Zhu, Yan, Li, Zhang, Wang, et~al.]{xun2025rtv}
Shuhang Xun, Sicheng Tao, Jungang Li, Yibo Shi, Zhixin Lin, Zhanhui Zhu, Yibo Yan, Hanqian Li, Linghao Zhang, Shikang Wang, et~al.
\newblock Rtv-bench: Benchmarking mllm continuous perception, understanding and reasoning through real-time video.
\newblock \emph{arXiv preprint arXiv:2505.02064}, 2025.

\bibitem[Yang et~al.(2025)Yang, He, Pan, Jiang, Deng, Yang, Lu, Yin, Rao, Zhu, Zhang, and Chen]{yang2025r1onevisionadvancinggeneralizedmultimodal}
Yi~Yang, Xiaoxuan He, Hongkun Pan, Xiyan Jiang, Yan Deng, Xingtao Yang, Haoyu Lu, Dacheng Yin, Fengyun Rao, Minfeng Zhu, Bo~Zhang, and Wei Chen.
\newblock R1-onevision: Advancing generalized multimodal reasoning through cross-modal formalization, 2025.
\newblock URL \url{https://arxiv.org/abs/2503.10615}.

\bibitem[Ye et~al.(2025)Ye, Zhang, Jiang, and Huang]{ye2025processsupervisedreinforcementlearningcode}
Yufan Ye, Ting Zhang, Wenbin Jiang, and Hua Huang.
\newblock Process-supervised reinforcement learning for code generation, 2025.
\newblock URL \url{https://arxiv.org/abs/2502.01715}.

\bibitem[Yu et~al.(2025)Yu, Lin, Zhao, Wei, Zhu, Wei, Sun, Ge, Zhang, Wang, et~al.]{yu2025unhackable}
En~Yu, Kangheng Lin, Liang Zhao, Yana Wei, Zining Zhu, Haoran Wei, Jianjian Sun, Zheng Ge, Xiangyu Zhang, Jingyu Wang, et~al.
\newblock Unhackable temporal rewarding for scalable video mllms.
\newblock \emph{arXiv preprint arXiv:2502.12081}, 2025.

\bibitem[Zhang et~al.(2025{\natexlab{a}})Zhang, Li, Cheng, Hu, Yuan, Chen, Leng, Jiang, Zhang, Li, et~al.]{zhang2025videollama}
Boqiang Zhang, Kehan Li, Zesen Cheng, Zhiqiang Hu, Yuqian Yuan, Guanzheng Chen, Sicong Leng, Yuming Jiang, Hang Zhang, Xin Li, et~al.
\newblock Videollama 3: Frontier multimodal foundation models for image and video understanding.
\newblock \emph{arXiv preprint arXiv:2501.13106}, 2025{\natexlab{a}}.

\bibitem[Zhang et~al.(2025{\natexlab{b}})Zhang, Wen, Wu, and Huang]{zhang2025tinyllava}
Xingjian Zhang, Siwei Wen, Wenjun Wu, and Lei Huang.
\newblock Tinyllava-video-r1: Towards smaller lmms for video reasoning.
\newblock \emph{arXiv preprint arXiv:2504.09641}, 2025{\natexlab{b}}.

\bibitem[Zhang et~al.(2023)Zhang, Liu, Tao, Chen, Fei, Che, and Qin]{vitcot}
Yongheng Zhang, Xu~Liu, Ruihan Tao, Qiguang Chen, Hao Fei, Wanxiang Che, and Libo Qin.
\newblock {ViTCoT}: Video-text interleaved chain-of-thought for boosting video understanding in large language models.
\newblock In \emph{Proceedings of the 31st ACM International Conference on Multimedia (ACM MM)}, 2023.

\end{thebibliography}
